\documentclass[runningheads]{llncs}

\usepackage{eccv}

\usepackage{eccvabbrv}
\usepackage[leftcaption]{sidecap}

\usepackage{graphicx}
\usepackage{booktabs}
\usepackage{ulem}
\usepackage[accsupp]{axessibility}  

\usepackage{color, colortbl,booktabs}
\definecolor{Gray}{gray}{0.9}

\usepackage{hyperref}

\usepackage{orcidlink}
\usepackage{algorithm}
\usepackage{algpseudocode}

\begin{document}

\title{AFreeCA: Annotation-Free Counting for All} 
\titlerunning{Annotation-Free Counting}

\author{Adriano D'Alessandro\orcidlink{0009-0004-1791-8843} \and
Ali Mahdavi-Amiri\orcidlink{0000-0002-4693-3565} \and
Ghassan Hamarneh\orcidlink{0000-0001-5040-7448}}

\authorrunning{A.~D'Alessandro et al.}

\institute{Simon Fraser University, Burnaby, Canada\\
\email{\{acdaless,amahdavi,hamarneh\}@sfu.ca}}

\definecolor{green}{rgb}{0, 0.5, 0}
\definecolor{orange}{rgb}{0.8, 0.6, 0.2}
\definecolor{red}{rgb}{1.0, 0.0, 0.0}
\definecolor{teal}{rgb}{0.0, 0.4, 0.4}
\definecolor{purple}{rgb}{0.65,0,0.65}
\definecolor{saffron}{rgb}{0.95,0.75,0.2}
\definecolor{turquoise}{rgb}{0.0,0.5,0.5}
\definecolor{black}{rgb}{0.0, 0.0, 0.0}
\definecolor{gray}{rgb}{0.5, 0.5, 0.5}

 \newcommand{\ali}[1]{\textcolor{black}{[Ali: #1]}}
 \newcommand{\am}[1]{{\color{black}#1}}
 \newcommand{\adrian}[1]{{\color{black}#1}}

\maketitle

\begin{abstract}
    Object counting methods typically rely on manually annotated datasets. The cost of creating such datasets has restricted the versatility of these networks to count objects from specific classes (such as humans or penguins), and counting objects from diverse categories remains a challenge. The availability of robust text-to-image latent diffusion models (LDMs) raises the question of whether these models can be utilized to generate counting datasets. \am{However, LDMs struggle to create images with an exact number of objects based solely on text prompts but they can be used to offer a dependable \textit{sorting} signal by adding and removing objects within an image. Leveraging this data, we initially introduce an unsupervised sorting methodology to learn object-related features that are subsequently refined and anchored for counting purposes using \adrian{counting data} generated by LDMs.} Further, we present a density classifier-guided method for dividing an image into patches containing objects that can be reliably counted. Consequently, we can generate counting data for any type of object and count them in an unsupervised manner. Our approach outperforms unsupervised and few-shot alternatives and is not restricted to specific object classes for which counting data is available.  Code available at: \href{https://github.com/adrian-dalessandro/AFreeCA}{github.com/adrian-dalessandro/AFreeCA}.
  \keywords{Object Counting \and Synthetic Data \and Feature Learning}
\end{abstract}

\section{Introduction}
\label{sec:intro}
\begin{figure}[t]
  \centering
  \includegraphics[width=\textwidth]{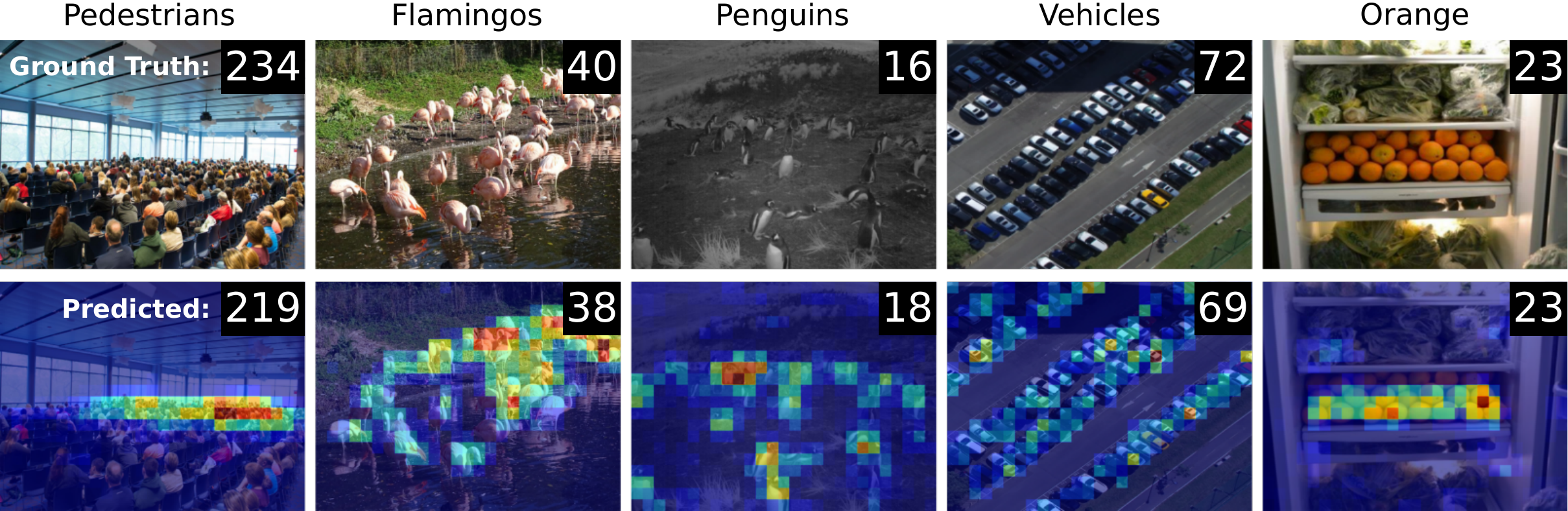}
  \caption{We propose a method which exploits synthetic counting data generated by Stable Diffusion. With this, we establish an \textit{annotator-free} method that produces accurate count maps without location-based supervision for a wide range of object categories.}
  \label{fig:many_examples}
\end{figure}

\begin{figure}[t]
  \centering
  \includegraphics[width=\textwidth]{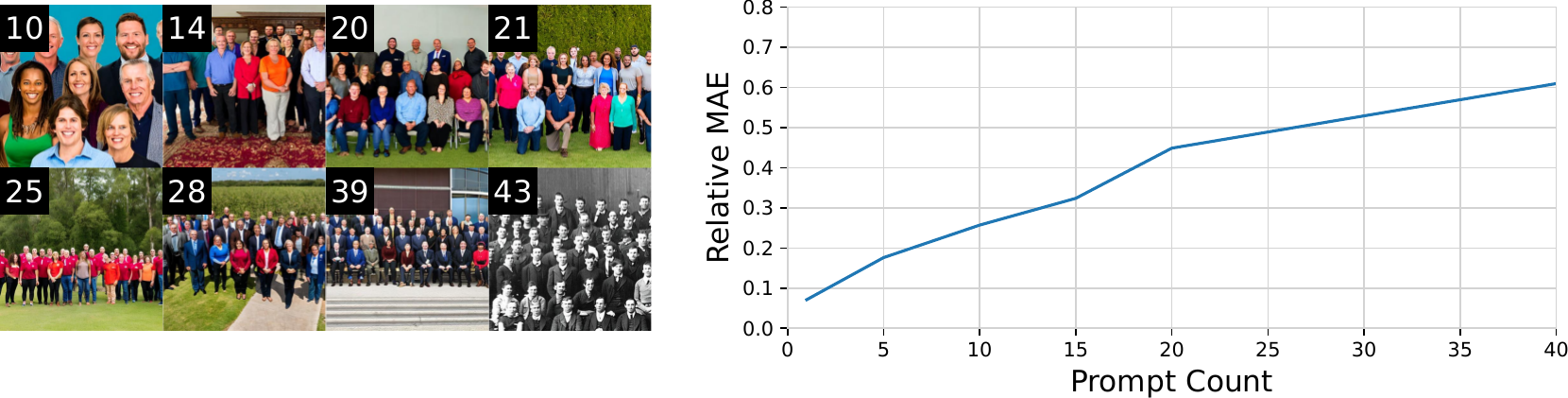}
  \caption{\textit{Left:} when given a prompt count of 20, Stable Diffusion outputs images with a similar \adrian{but often incorrect object count}. \textit{Right:} as the prompt count increases, the relative error between the true underlying count and the prompt count increases.}
  \label{fig:noisy_count}
\end{figure}

\begin{figure}[t]
  \centering
  \includegraphics[width=\textwidth]{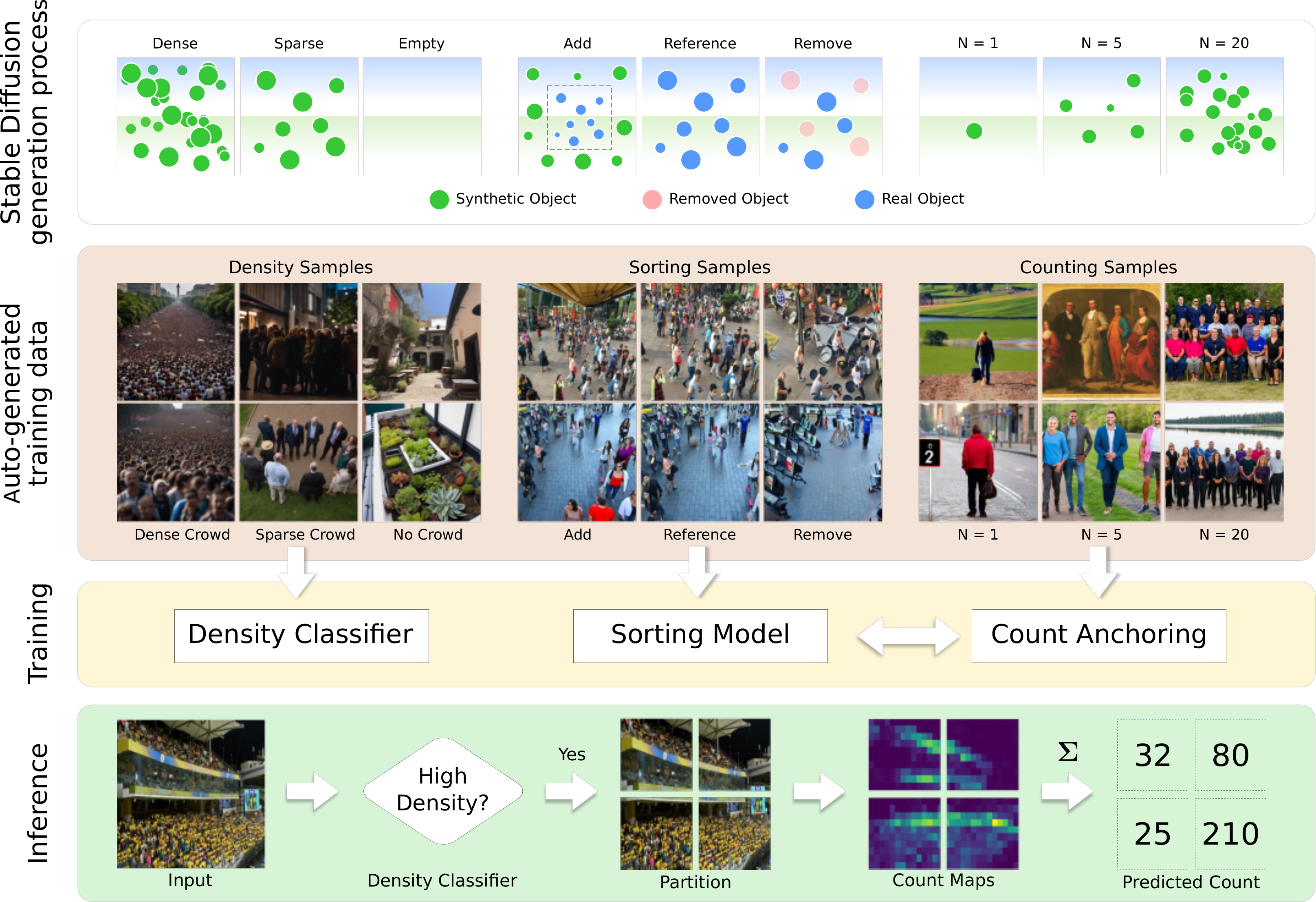}
  \caption{\textbf{Workflow.} Our framework uses simple prompts to create synthetic data for training a sorting model, a density classifier, and a count anchoring network. These elements are combined into a model which can accurately count diverse object categories even within dense images by subdividing them into smaller, more manageable areas.}
  \label{fig:method_teaser}
\end{figure}

Object counting is \am{an important} task in computer vision, with diverse applications such as crowd analysis~\cite{sindagi2020jhu, zhang2016single, idrees2018composition}, wildlife monitoring~\cite{Arteta16}, and traffic analysis~\cite{Hsieh_2017_ICCV}. These tasks, historically reliant on fully-supervised methods, demand detailed annotations that are labor-intensive to collect. For instance, the annotation of 5,109 images for a crowd counting dataset necessitated 3,000 hours, averaging 35 minutes per image~\cite{gao2020nwpu}. Although semi-supervised approaches have sought to reduce this burden by requiring annotations for only a subset of the data, they still demand hundreds of hours of labor. In response, research into unsupervised counting methods has aimed to eliminate the annotation burden entirely. Despite these efforts, such methods remain challenging to develop \am{and the existing methods are limited to crowd counting~\cite{css_ccnn_2022, Liang_2023_CVPR}.}
\am{In this paper, using text-to-image latent diffusion models (LDMs), we propose an \textit{unsupervised} counting method that can be applied to a \textit{wide variety} of objects as in \cref{fig:many_examples}.}

Recently, advancements in text-to-image LDMs, such as Stable Diffusion~\cite{rombach2022high}, have presented novel opportunities to address \am{various} challenges. These methods excel at producing synthetic images \adrian{with a high degree of realism} that accurately reflect the contents of text-based prompts. This capability enables the generation of labeled synthetic images automatically by integrating label information into the prompts. Recent works have highlighted the significant potential of this strategy for zero-shot recognition, few-shot recognition, and model pre-training~\cite{Shipard_2023_CVPR,he2023is}. However, these strategies have only focused on object recognition, raising the question of whether they can be applied to object counting.

\am{A straightforward approach for extending the use of LDMs to object counting might involve prompting the network to produce images using explicit quantity labels, such as ``\textit{An image with 20 people}'' with ``20'' used as the prompt count label \adrian{as seen in \cref{fig:noisy_count} (Left)}. However, as previous work has also indicated, the text encoders used in LDMs have a limited understanding of quantity~\cite{Paiss_2023_ICCV}.} \am{To demonstrate this, we annotated 40 synthetic samples per-category across different crowd count categories and analyzed the discrepancy between prompted and actual counts, termed label noise. \cref{fig:noisy_count} (Right) shows that the relative discrepancy increases with the magnitude of the prompt count label, \adrian{suggesting that} synthetic data is unreliable for images with high counts.}
\adrian{To learn robust count-related features given this label noise we train a sorting network on a dataset comprised of image triplets, ordered by their known count ranking. These triplets are generated by synthetically adding to or removing objects from real images using LDMs, enabling the network to capture high-quality counting features. While these features do not directly correspond to specific count numbers, we anchor them using synthetic counting data. To accomplish this, we fine-tune only a linear layer on-top of the sorting network, ensuring the stability of the learned features~\cite{kumar2022finetuning, trivedi2023a}. Further, given the diminished reliability of the synthetic data for high object counts, we adopt an approach at inference-time to divide dense regions into smaller patches. This partitioning strategy not only aligns the patch with the more reliable counting range of our network but it also helps in preserving high-frequency details by sourcing higher-resolution patches directly from the original images. Our method's workflow is detailed in \cref{fig:method_teaser}.}
In summary, our contributions are listed in the following.
\am{
\begin{itemize}
\item To the best of our knowledge, we propose the \textit{first unsupervised} counting method that can be used for a \textit{variety of object categories}.  
\item\am{We demonstrate that latent diffusion models are able to produce valuable counting and sorting data across diverse object categories. We introduce a method to leverage this data, initially focusing on learning to sort followed by a fine tuning strategy to learn counting.}
\item To maximize the utilization of image resolution and improve performance on highly dense images, we introduce a method that dynamically partitions dense images by leveraging guidance from a density classification network.
\item We surpass SOTA unsupervised and zero-shot crowd counting methods on several crowd counting benchmark datasets. We also provide various ablation studies to justify our design.
\end{itemize}}

\section{Related Work}
\paragraph{Fully-Supervised Counting.} 
Object counting has advanced through density map learning, using Gaussian kernels convolved with dot map annotations~\cite{zhang2016single, lempitsky2010learning,li2018csrnet}. There are a wide range of diverse approaches to leveraging these annotations. GauNet~\cite{Cheng_2022_CVPR} employs locally connected Gaussian kernels for density map generation, introducing low-rank approximation with translation invariance. GLoss~\cite{wan2021generalized} reframes density map estimation as an unbalanced optimal transport problem, proposing a specialized loss function. And, AMSNet~\cite{hu2020count} uses neural architecture search and a multi-scale loss to incorporate structural information.
\paragraph{Unsupervised Counting.} In our work, we define unsupervised counting as any method that eliminates the need for new annotations, thus reducing the annotation burden for object counting problems. This definition includes methods that utilize foundation models. CSS-CCNN~\cite{css_ccnn_2022} introduces an unsupervised approach using self-supervised pre-training with image rotation and Sinkhorn matching, requiring prior knowledge of maximum crowd count and power law distribution parameters. CrowdCLIP~\cite{Liang_2023_CVPR} leverages features from CLIP, a pre-trained language-image model, through multi-modal ranking with text prompts. They use a filtering strategy to select crowd-containing patches and match features to crowd count categories represented in text prompts.
\paragraph{Generative Models.} Stable Diffusion (SD)~\cite{rombach2022high} is a generative model for image synthesis based on latent diffusion models. SD generates images through a multi-step process. Initially, a real image is encoded via a variational autoencoder, yielding a compressed representation in a lower-dimensional latent space. Gaussian noise is then iteratively applied to these features. To restore meaningful features, SD employs the reverse diffusion process, an iterative denoising mechanism. Finally, the denoised features are decoded to synthesize a new image. SD can be conditioned on text embeddings, typically from a text-encoder like CLIP~\cite{radford2021learning}, enabling image generation guided by user-defined text prompts.

\section{Methodology} \label{sec:methods}
To count objects in images from an \textit{arbitrary} category, we propose using latent diffusion models (LDMs) to synthesize data. However, LDMs often misinterpret quantity in prompts, leading to discrepancies between the intended and actual object counts in generated images, known as \textit{label noise}. This issue worsens with higher prompt counts, highlighting two key problems: the existence of a count threshold beyond which generated data becomes unreliable due to excessive label noise, and the diminished reliability of learned features for dense regions, increasing the risk of the model overfitting to noise. Our methodology addresses these problems by combining several strategies explained in the following.

First, we generate highly reliable synthetic sorting data by adding and removing objects from an image using Stable Diffusion (SD)~\cite{rombach2022high} (Fig.~\ref{fig:method_teaser}). This process produces triplets of images ranked based on object count that can be used to learn robust counting features through training a \textit{sorting network}~\cite{liu2018leveraging, daless2023, Liang_2023_CVPR}.
Following this, we use SD to generate \textit{noisy} synthetic counting data. This data is then used to fine-tune a counting network built upon our sorting network. By using the count data as anchors within the sorting features, we can preserve the reliability of the object quantity features obtained by the sorting network while establishing a correspondence between the features and the count.

As the data generated by SD tends to be more accurate for smaller counts, there could be a decline in performance when dealing with images containing a large number of objects.
We extend the model's reach to a wider range of object counts by recognizing that any image can be partitioned into sub-regions with fewer objects.
However, over-partitioning sparse regions can introduce noise due to boundary artifacts, double counting, etc. Thus, we need a method to identify dense regions within an image. As a result, we generate synthetic crowd density classification data and then fine-tune a density guidance network on top of our pre-trained sorting model to identify dense image regions. At the inference stage, this density guidance is instrumental in identifying image sections that might exceed the counting model's reliable counting range. These identified regions are then processed at a higher resolution (when available) to ensure accurate counting. By retrieving higher-resolution patches from the original image, we address issues related to feature loss caused by resizing images to a fixed size before feeding them into a network. Therefore, dividing an image into patches leads to more precise counting estimations.

\subsection{Learning to Sort}
\label{sec:learning_to_sort}
\subsubsection{Generating Synthetic Sorting Data}
We generate a dataset of \adrian{images sorted based on their object counts} by modifying reference images to either add or remove objects, using a LDM for image-to-image synthesis or out-painting. Unlike previous approaches that focus on intra-image ranking for counting~\cite{liu2018leveraging}, our strategy can add objects and also remove objects while preserving the scene's perspective and original features. This broadens the diversity and range of object counts in the resulting image triplets.
The process begins with a base dataset of real or \am{generated} reference images, represented as \adrian{$\mathcal{D}^{\text{ref}} = \{x^{\text{ref}}_i\}^{N_{\text{ref}}}$}, where \( x^{\text{ref}}_i \) denotes an image with an unknown number of objects \( c^{\text{ref}}_i \), and \( N_{\text{ref}} \) is the total number of such images.
To synthesize images with fewer objects, we apply Stable Diffusion ($SD$):
\begin{equation}
x^{\text{syn}-}_i = SD(x^{\text{ref}}_i, t_{\text{p}}, t_{\text{n}})
\end{equation}

\noindent where \( x^{\text{syn}-}_i \) is the resultant image with fewer objects, guided by a text prompt \( t_{\text{p}} \) and,  to prevent the addition of targeted objects, a negative prompt \( t_{\text{n}} \). As an example, we set \( t_{\text{p}} \) to ``an empty place'' and \( t_{\text{n}} \) to ``pedestrians, humans, people, crowds'' for crowd counting problems. Due to the stochastic nature of the image generation process, $x_i^{\texttt{syn}-}$ contains an unknown number of objects $c^{\texttt{syn}}_i$. In the supplementary material, we empirically verify that the relationship, $c^{\texttt{ref}}_i \geq c^{\texttt{syn}}_i,$
holds in 99\% of cases. 

For adding objects, we engage in a similar process using outpainting:
\begin{equation}
x^{\text{syn}+}_i = SD(x^{\text{ref}}_i, \mathcal{M}, t_{\text{p}}, t_{\text{n}})
\end{equation}
\noindent where \( \mathcal{M} \) is a mask indicating where to outpaint and, thus, add new objects and is set to a thick band around the perimeter of the image $x^{\text{ref}}_i$, which is 1/3rd of the image size. This approach effectively increases the object count and scene density. 
We transform each image \( x_i^{\text{ref}} \) into four augmented versions with increased object counts \(\{ x^{\text{syn}+}_{ij}\}_{j=1}^4 \) and four with decreased counts \(\{ x^{\text{syn}-}_{ik}\}_{k=1}^4 \), from which the 16 ordered triplets $\{x^{\text{syn}-}_{ij} \ge  x_i^{\text{ref}} \ge x^{\text{syn}+}_{ik}\}_{j,k=1}^4$ are generated.

\subsubsection{Pre-Training}
For a given triplet of images $X = \{x^{\text{syn}-}, x^{ref}, x^{\text{syn}+}\}$ with rank labels $Y = \{0, 1, 2\}$, we generate a similarity matrix
$[\mathcal{S}^{y}]_{i,j}=-|y_i-y_j|$.
We encode $X$ using network $f_\theta$ to produce feature vectors $Z = \{z^{syn-}, z^{ref}, z^{syn+}\}$; $z_i \in \mathbb{R}^{2048}$.
We further utilize a sorting head $v_\Theta$ to produce a continuous valued output $\hat{y}_i$ from $z_i$ as an estimate of the rank of $x_i$. We then calculate a second similarity matrix $[\mathcal{S}^{z}]_{i,j}$ using the cosine similarity between $z_i$ and $z_j$. $\mathcal{S}^{z}$ and $\mathcal{S}^{y}$
encode the relational structure between examples and are used to align the feature space and label space with the ground truth ordering by minimizing the following losses (this is akin to RankSim~\cite{gong2022ranksim} with the key difference that we use count sorting labels instead of ground truth values like they do):
\begin{equation}
\ell^{y}_{sort} = \sum_i^{3}(\mathbf{rk}(S^{y}_i) - \mathbf{rk}(S^{\hat{y}}_{i}))^2\text{~~~~~~~}\ell^{z}_{sort} = \sum_i^{3}(\mathbf{rk}(S^{y}_i) - \mathbf{rk}(S^{z}_{i}))^2,
\end{equation}
\noindent where $S^{\hat{y}}_i$ is the predicted similarity matrix, $S^{*}_i$ is $i$-th row of $S^{*}$, and $\mathbf{rk}$ is a non-differentiable ranking function.  The total sorting loss is then defined as:
\begin{equation}
\mathcal{L}_{sort} = \ell^{y}_{sort} + \lambda \ell^{z}_{sort},
\end{equation}

\noindent where $\lambda$ is set to $5.0$. To optimize $\mathbf{rk}$, we follow the same strategy as in RankSim~\cite{gong2022ranksim} and apply a blackbox combinatorial solver~\cite{2020Differentiation}. 

\subsection{Learning to Count from Synthetic Data}
\label{sec:learning_to_count}

Our approach generates synthetic images with approximate object counts using $SD$ for text-to-image synthesis: 
\begin{equation}
\{x^{\text{s}}_i, c^{\text{p}}_i\} = SD(t_{\text{p}}, t_{\text{n}}),
\end{equation}
where $c^{\text{p}}_i$ is the object count from the prompt, and $x^{\text{s}}_i$ are the generated images. This yields a dataset of synthetic examples \adrian{$\mathcal{D}^{\text{s}}_{\text{cnt}} = \{x^{\text{s}}_i, c^{\text{p}}_i\}^{N_{\text{cnt}}}.$}
We use simple prompts to ensure accurate image generation, and to avoid complex prompt engineering. For crowd counting, prompts are straightforward, e.g., ``20 people.'' Our experiments involve generating images for a wide range of counts from 1 to 1000, which we elaborate on in the supplementary material. We generate 150 images for each prompt category, plus 800 images with zero objects generated by random sampling scene category names from the Places365~\cite{zhou2017places} dataset for prompts and explicitly excluding objects using negative prompts. 
\begin{figure}[tb]
  \centering
  \includegraphics[width=\textwidth]{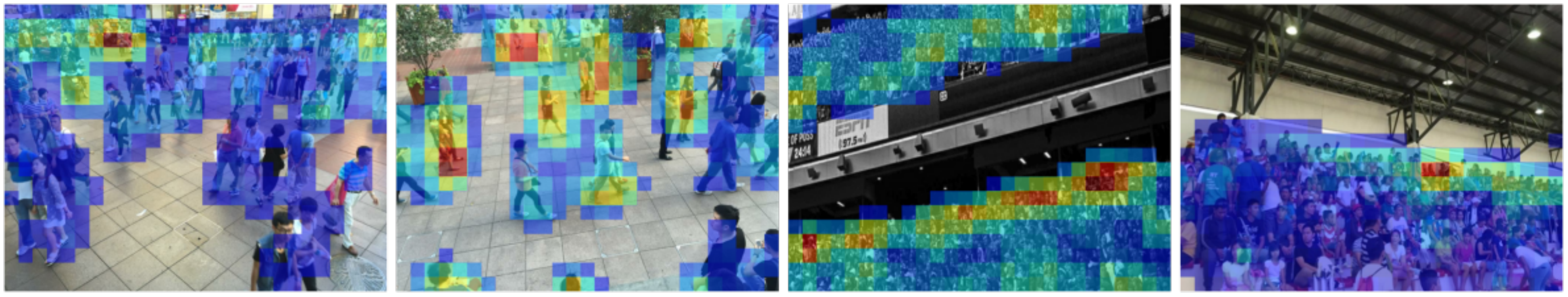}
  \caption{\textbf{Sorting Features}. We calculate the channel-wise mean of the features produced by the sorting network to demonstrate where the network is active. The network appears to focus on the object of interest across a wide range of crowd densities.}
  \label{fig:synth_sort_gen}
\end{figure}
For training our counting network with the synthetic dataset $\mathcal{D}^{\texttt{s}}_{\texttt{cnt}}$ and the pre-trained sorting network $f_\theta$, we adopt a strategy to preserve the integrity of $f_\theta$ features. Previous works have indicated that fine-tuning an entire network distorts the pre-trained features of that network and degrades performance on out-of-distribution data~\cite{kumar2022finetuning, trivedi2023a}. In \cref{fig:synth_sort_gen}, we observe that the feature maps produced by $f_\theta$ focus on the image regions where objects exist, which is a property that we would like to preserve.
Given that we are training our counting network using synthetic data, and we are evaluating on real data, to avoid the potential for any feature distortion \am{we only fine-tune a linear layer atop the pre-trained sorting network to anchor its features to actual count values.}

To further ensure that our counting network is as accurate as possible, we \am{automatically} filter out outliers from the synthetic dataset \am{using a feature space analysis approach, which is effective for mitigating the impacts of label noise~\cite{NEURIPS2020_f4e3ce3e, pmlr-v162-zhu22a, Lee_2018_CVPR}.} We adopt a simplified version of CleanNet\cite{Lee_2018_CVPR}, which calculates class prototypes as well as the similarity between samples and prototypes to detect noisy samples. We compute features $z_i$ for each image $x^{s}_i$ using $f_\theta$ and then create a prototype vector $z^{c}_{\mu}$ for each prompt count category $c$:
\begin{equation}
z^{c}_{\mu} = \frac{1}{N} \sum_{i=1}^{N} z^{c}_i.
\end{equation}
We then filter out synthetic images whose features align closer with a different category's prototype than their own. Finally, we train a counting network, $g_\Phi$ \am{optimizing a mean squared error}:
\begin{equation}
   \mathcal{L}_{count} = \frac{1}{N_{cnt}} \sum_{ (x_i^{\texttt{s}}, c^{\texttt{p}}_i) \in \mathcal{D}^{s}_{cnt}} (g_{\Phi} \circ
f_\theta(x_i^{\texttt{s}}) -  c^{\texttt{p}}_i)^2. 
\end{equation}

\subsection{Crowd Density Classification}
\am{Given that \adrian{our counting networks is more reliable for} smaller counts, we partition \adrian{dense images} into \adrian{patches} with lower counts to enhance estimation accuracy in each \adrian{patch}. This partitioning process is directed by a classifier that discerns between dense and sparse images. Given that dense image regions may be poorly represented by our counting model, a dataset is necessary for training such a classifier.}
We use stable diffusion to produce a classification dataset of synthetic images with various crowd densities \adrian{$\mathcal{D}_{dense}^{s} = \{x_i^{s}, y_i^{den}\}^{N_{den}}$}, where $x_i^{s}$ is the synthetic image, $y_i^{den}$ is the density category recorded in the prompt, and $N_{den}$ is the size of the dataset. As highlighted in \cref{fig:method_teaser}, we categorize densities into three groups: no crowd, sparse crowd, and dense crowd, embedding these categories directly into the generation prompts to create representative images for each. We re-purpose the zero-object examples from the earlier noisy count dataset for our "no crowd" category.
\am{We then use this synthetic dataset to train our classifier by fine-tuning a density classification layer, $h_{\phi}$, on top of pre-trained sorting network. The loss function for this step is:}
\begin{equation}
\mathcal{L}_{dense} = \frac{1}{N_{den}} \sum_{ (x_i^{\texttt{s}}, y^{\texttt{den}}_i) \in \mathcal{D}^{s}_{den}} \ell_{CE}(h_{\phi} \circ
f_\theta(x_i^{\texttt{s}}), y^{\texttt{den}}_i) ,\end{equation} where $\ell_{CE}$ is the cross-entropy loss.

\begin{figure}[tb]
  \centering
  \includegraphics[width=\textwidth]{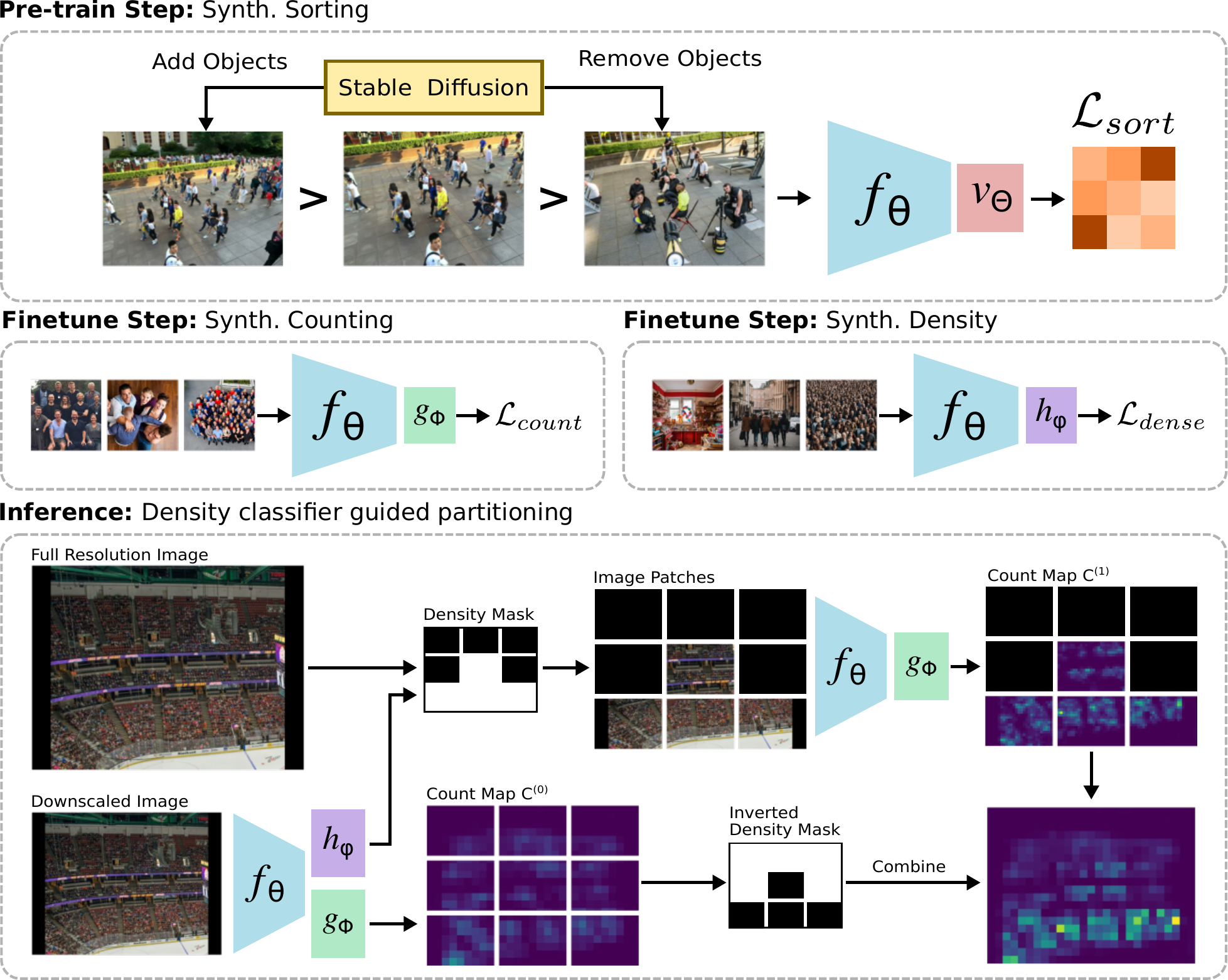}
  \caption{\textbf{Methodology}. Our strategy involves three distinct steps supported by a synthetic training signal extracted from stable diffusion. The pre-training step sorts synthetic and real images to learn high quality object quantity features from the source distribution. The finetuning step utilizes the pre-trained features and synthetic data to train a counting head and a density classification head. Finally, the density classifier guides inference by partitioning dense images so that there are fewer objects per image.}
  \label{fig:methodology}
\end{figure}

\subsection{Density Classifier Guided Partitioning (DCGP)}
We previously argued that by partitioning images into sub-region patches, each patch will have fewer objects and be more likely to fall within the accurate range of our counting model. However, images often contain groups of objects at a variety of sizes and crowd densities due to the perspective of the scene. Given this, we propose a technique called density classifier guided partitioning (DCGP), \adrian{which treats image patches differently based on their estimated density}.

\am{
For each image, we produce a count map to approximate the number of objects within the respective region (see Fig. \ref{fig:methodology}, bottom). In cases where a region exhibits sparsity according to our density classifier, like the top-right quadrant of the count map depicted in Fig. \ref{fig:methodology}, the original count maps can be directly employed to estimate the count within that region. Conversely, if a count map corresponds to a dense region, the spatial patch associated with that region, potentially at a higher resolution, is extracted from the input image. This patch is then forwarded to our counting network to obtain a more precise estimate, considering the smaller number of objects within the patch, enabling the counting network to perform more effectively in estimation. This hybrid resolution counting estimation is eventually combined to find the final count. 
}

\adrian{Formally, to obtain a count map we adapt our pre-trained sorting network $f_\theta$, omitting the global average pooling layer. This modified network produces feature maps $z \in \mathbb{R}^{H \times W \times 2048}$, where $H$ is the height of the feature map and $W$ is the width. We calculate the regional counts for each feature map element as follows:
\begin{equation}c^{(0)}_{ij} = g_\Phi(z_{ij}/(H \cdot W)),\end{equation}
where $c^{(0)}_{ij}$ denotes the count for the $(i,j)$-th element, providing a comprehensive count map fir the image. We further classify each region's density using $f_\theta$ and a density classifier $h_\phi$ to generate a density map $D$, where $D_{ij}$ represents the density class for the $(i,j)$-th region in $z$. This allows us to estimate which regions in the whole image are dense and require further processing.

Finally, we partition a high resolution version of a given image into a $3 \times 3$ grid of image patches. For sparse patches, we directly sum counts from $c^{(0)}$. For dense patches, we process them through the counting network for a precise patch count estimate $c^{(1)}$, summing these to achieve the image's final count.}

\begin{table*}[t]
\small
\caption{\textbf{Crowd Counting Performance.} We evaluate the performance of our method on the test set of crowd counting benchmark datasets. \am{It is evident that our method outperforms other unsupervised and zero-shot techniques. }}
\centering
\begin{tabular}{llcccccccc}
\hline
\multicolumn{2}{c}{}&  \multicolumn{2}{c}{SHB}  & \multicolumn{2}{c}{JHU} &  \multicolumn{2}{c}{SHA} &  \multicolumn{2}{c}{QNRF} \\
\cline{3-10}
Method & Type & \multicolumn{1}{c}{MAE} & \multicolumn{1}{c}{MSE} & MAE & MSE & MAE & MSE & MAE & MSE \\
\hline
ADSCNet~\cite{Bai_2020_CVPR}
       & Full supervised & 6.4 & 11.3 &  - & -  & 55.4 & 97.7 & 71.3 & 132.5  \\
GLoss~\cite{wan2021generalized} 
       & Full supervised & 7.3 & 11.7 & 59.9 & 259.5 & 61.3 & 95.4  & 84.3 & 147.5\\
GauNet~\cite{cheng2022rethinking}
       & Full supervised & 6.2 & 9.9 & 58.2 & 245.1 &  54.8 & 89.1 & 81.6 & 153.7 \\
\hline
CLIP-Count~\cite{clip_count_2023}
       & Zero-Shot & 45.7 & 77.4 & - & - & 192.6 & 308.4 & - & - \\
CSS-CCNN++~\cite{css_ccnn_2022} 
       & \textcolor{BrickRed}{Unsupervised} & - & - & 197.9 & 611.9 & 195.6 & 293.2  & 414.0 & 652.1 \\
CrowdCLIP~\cite{Liang_2023_CVPR}
       & \textcolor{BrickRed}{Unsupervised} & 69.3 & 85.8 & 213.7 & 576.1 & \textbf{146.1} & 236.3  & 283.3 & 488.7 \\
\rowcolor{Gray} Ours 
       & \textcolor{BrickRed}{Unsupervised} & \textbf{35.0} & \textbf{50.7} & \textbf{173.8} & \textbf{519.4} & 152.7 & \textbf{219.0} & \textbf{283.1} & \textbf{453.2} \\

\hline
\end{tabular}
\label{tab:crowd_performance}
\end{table*}

\begin{table*}[t]
\small
\caption{\textbf{Object Counting Performance.} Our method outperforms few-shot and zero-shot methods on the CARPK dataset.}
\label{tab:object_performance}
\centering
\begin{tabular}{llccc}
\hline
\multicolumn{2}{c}{} & \multicolumn{2}{c}{CARPK}\\
\cline{3-4}
Method & Type &   MAE & MSE  \\
\hline

FamNet~\cite{famnetcvpr2021}
       & Few-shot &  28.84 & 44.47  \\
BMNet+~\cite{min2022bmnet}
       & Few-shot & 10.44 & 13.77  \\
CLIP-Count~\cite{clip_count_2023}
       & Zero-shot & 11.96 & 16.61          \\
\rowcolor{Gray} Ours 
       & \textcolor{BrickRed}{Unsupervised} & \textbf{9.35} & \textbf{12.29}\\

\hline
\end{tabular}

\end{table*}
\begin{figure}[ht]
  \centering
  \includegraphics[width=\textwidth]{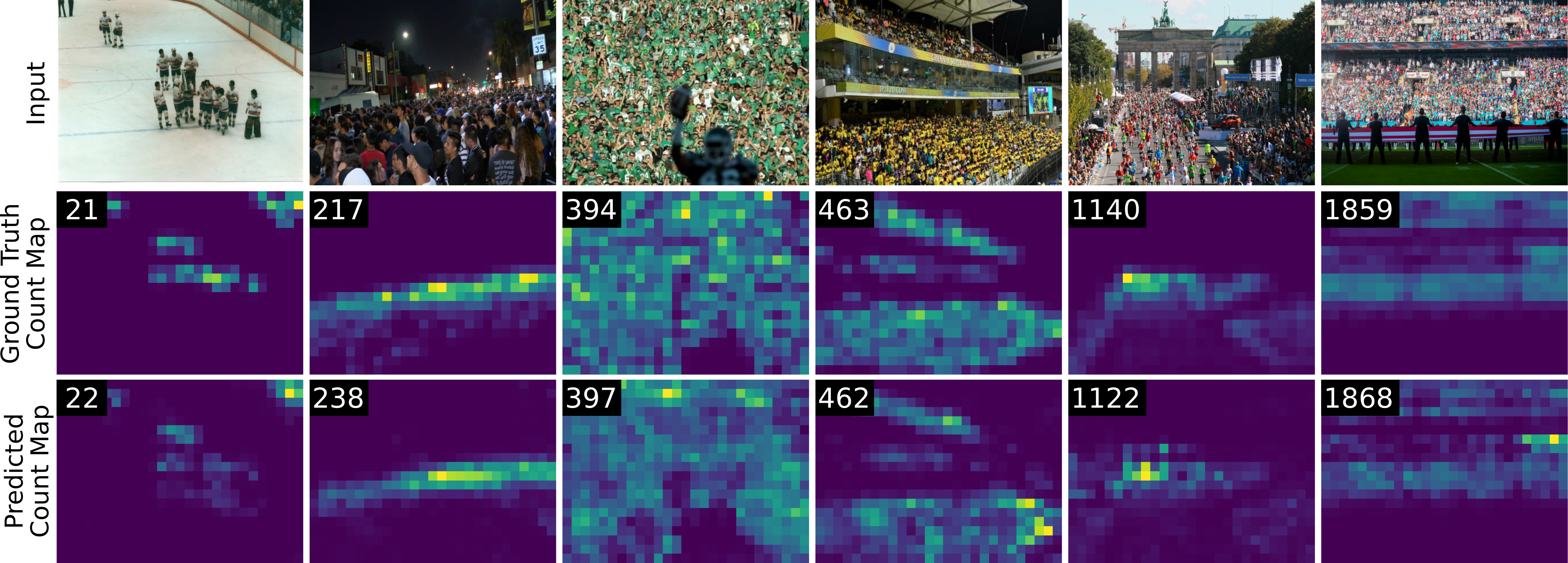}
  \caption{\textbf{Qualitative Crowd Counting}. We compare the count maps produced by our model to the ground truth count maps. The count maps are annotated with the ground truth counts and predicted counts. Our method accurately localizes crowds across a range of crowd densities, without any location-based supervision.}
  \label{fig:qual_results}
\end{figure}

\section{Experiments \& Results} 

\begin{figure}[th]
  \centering
  \includegraphics[width=\textwidth]{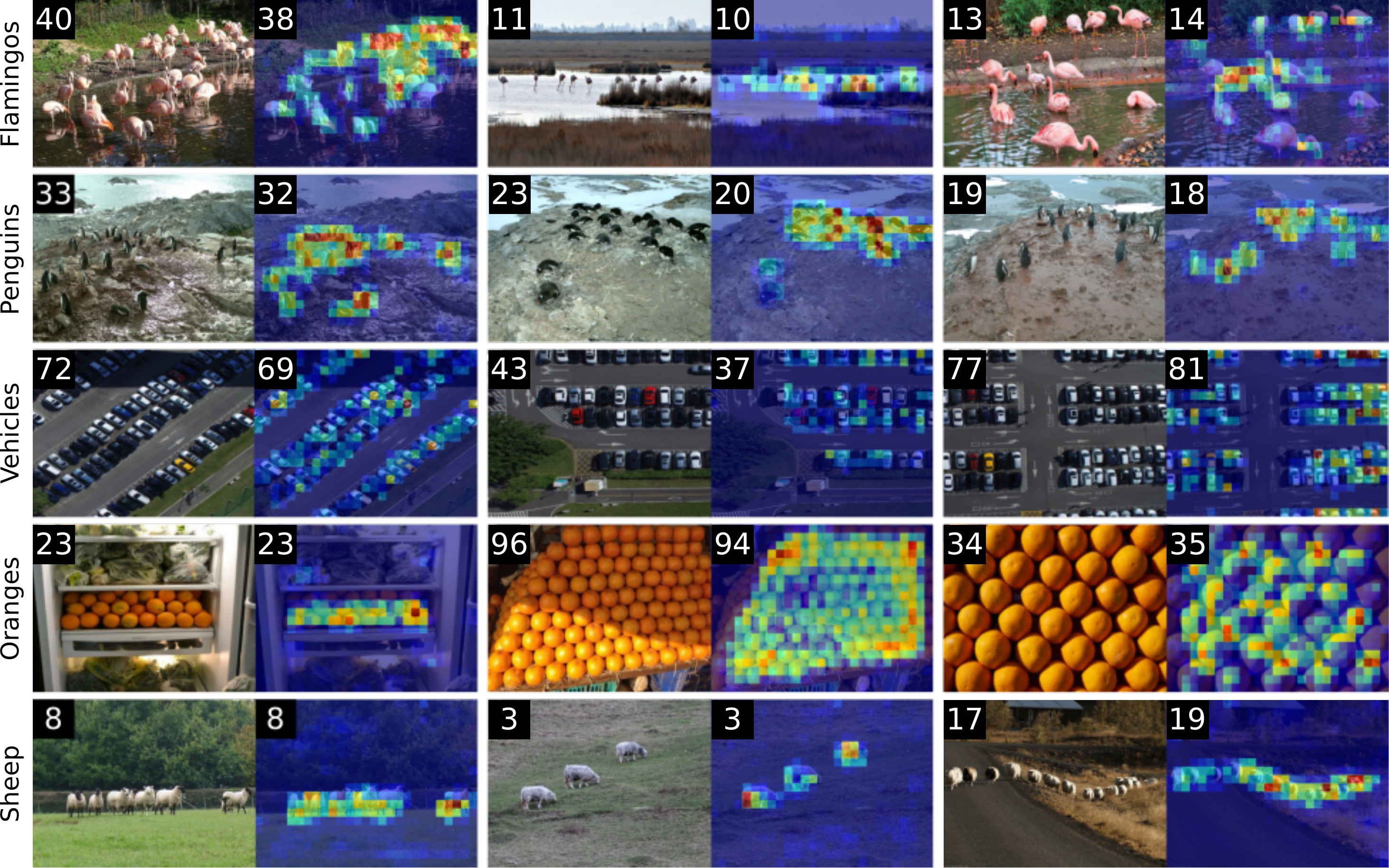}
  \caption{\textbf{Qualitative Object Counting}. We evaluate our method across a diverse range of object categories and demonstrate its proficiency in accurately predicting object counts and also generating precise count maps for each evaluated category.}
  \label{fig:qual_results_obj}
\end{figure}
\begin{table}[t]
  \caption{\textbf{Partitioning Strategy.} We explore the impact of the image partitioning rate for naive partitioning and density classifier guidance.}
  \label{tab:partitioning_abl}
  \centering
    \begin{tabular}{llcccccccc}
    \hline
    \multicolumn{2}{c}{}&  \multicolumn{2}{c}{SHB}  & \multicolumn{2}{c}{JHU} &  \multicolumn{2}{c}{SHA} &  \multicolumn{2}{c}{QNRF} \\
    \cline{3-10}
    Strategy & Rate & \multicolumn{1}{c}{MAE} & \multicolumn{1}{c}{MSE} & MAE & MSE & MAE & MSE & MAE & MSE \\
    \hline
    Fixed & 1x1 & 42.1 &  68.1 & 203.6  & 649.6 & 196.6 & 321.4 & 392.5 & 707.1 \\
    Fixed & 2x2 & 51.6 &  62.1 & 202.4  & 531.0 & 165.7 & 258.2 & 314.5  & 484.8 \\
    Fixed & 3x3 & 47.1 &  57.1 & 219.5 & 501.8 & 186.9 &  278.8 & 425.6& 559.9 \\
    Simple GP & 1x1$\rightarrow$2x2 & 42.1 &  68.1 & 181.7  & 541.9 & 163.1 & 255.9 & 299.9  & 487.7 \\
    Simple GP & 1x1$\rightarrow$3x3 & 42.1 &  68.1 & 173.9  & \textbf{495.3} & 155.7 & 225.4 & 309.5  & 480.2 \\
    DCGP & 1x1$\rightarrow$2x2 & 38.5 &  61.0 & 182.8 & 562.9 & 156.9 & 230.8 & 324.0  & 541.7 \\
    DCGP & 1x1$\rightarrow$3x3 & \textbf{35.0}  & \textbf{50.7} & \textbf{173.8}  & 519.4 & \textbf{152.7} & \textbf{219.0} & \textbf{283.1} & \textbf{453.2} \\
  \bottomrule
  \end{tabular}
\end{table}
\begin{table}[h!]
  \caption{\textbf{Pre-training Ablation Study.} We explore the impact of different pre-training strategies on the final performance of the network}
  \label{tab:pre_training_abl}
  \centering
    \begin{tabular}{lcccccccc}
    \hline
    \multicolumn{1}{c}{}&  \multicolumn{2}{c}{SHB}  & \multicolumn{2}{c}{JHU} &  \multicolumn{2}{c}{SHA} &  \multicolumn{2}{c}{QNRF} \\
    \cline{2-9}
    Pre-training & \multicolumn{1}{c}{MAE} & \multicolumn{1}{c}{MSE} & MAE & MSE & MAE & MSE & MAE & MSE \\
    \hline
    ImageNet & 68.1 & 104.7 &  304.9 & 654.9 & 316.2 & 410.0 & 423.1 & 655.9  \\
    Intra-Image Rank & 52.8 & 76.3 &  256.4 & 610.4  & 186.6 & 291.2 & 380.0 &  694.9  \\
    $\ell^y_{sort}$ only & 38.2 & 58.0 &  183.9 & 551.1  & 188.4 & 293.5 & 359.3 & 579.1  \\
    \rowcolor{Gray} $\ell^y_{sort} + \ell^z_{sort}$ (Ours) & \textbf{35.0} & \textbf{50.7} & \textbf{173.8} & \textbf{519.4} & \textbf{152.7} & \textbf{219.0} & \textbf{283.1} & \textbf{453.2} \\ 
  \bottomrule
  \end{tabular}
\end{table}
\begin{table}[t]
  \caption{\textbf{Counting Network Ablation Study.} We explore the performance of different strategies when training the counting network with synthetic counting data.}
  \label{tab:layer_retraining}
  \centering
    \begin{tabular}{lcccccccc}
    \hline
    \multicolumn{1}{c}{}&  \multicolumn{2}{c}{SHB}  & \multicolumn{2}{c}{JHU} &  \multicolumn{2}{c}{SHA} &  \multicolumn{2}{c}{QNRF} \\
    \cline{2-9}
    Strategy & \multicolumn{1}{c}{MAE} & \multicolumn{1}{c}{MSE} & MAE & MSE & MAE & MSE & MAE & MSE \\
    \hline
    $\mathcal{L}_{count}$ only train &  122.9 & 154.6  & 318.5 & 785.7 & 408.8 & 531.8 & 711.2 & 1027.6 \\
    Full Network Finetune & 43.9 & 74.6 &  292.4 & 749.0  & 340.8 & 462.5 & 664.8 & 965.2  \\
    \rowcolor{Gray}Last Layer Finetune & \textbf{35.0}& \textbf{50.7} &  \textbf{173.8} & \textbf{519.4}  & \textbf{157.7} & \textbf{219.0} & \textbf{283.1} & \textbf{453.2}  \\
  \bottomrule
  \end{tabular}
\end{table}
\begin{table}[ht]
  \caption{\textbf{High Resolution Partitioning.} We explore the performance implications of selecting partitions from a high-resolution image compared to a low resolution image.}
  \label{tab:ablate_resolution}
  \centering
    \begin{tabular}{lcccccccc}
    \hline
    \multicolumn{1}{c}{}&  \multicolumn{2}{c}{SHB}  & \multicolumn{2}{c}{JHU} &  \multicolumn{2}{c}{SHA} &  \multicolumn{2}{c}{QNRF} \\
    \cline{2-9}
    Strategy & \multicolumn{1}{c}{MAE} & \multicolumn{1}{c}{MSE} & MAE & MSE & MAE & MSE & MAE & MSE \\
    \hline
    Low Resolution & 35.6 & 55.0 &  180.2 & 563.8  & 154.4 & 223.8 & 317.6 & 552.8  \\
    \rowcolor{Gray} Full Resolution & \textbf{35.0} & \textbf{50.7} &  \textbf{173.8} & \textbf{519.4}  & \textbf{152.7} & \textbf{219.0} & \textbf{283.1} & \textbf{453.2}  \\
  \bottomrule
  \end{tabular}
\end{table}

\subsection{Performance}
\subsubsection{Datasets.}
We benchmark on 4 crowd counting datasets: ShanghaiTechA (SHA)~\cite{zhang2016single}, ShanghaiTechB (SHB)~\cite{zhang2016single}, UCF-QNRF~\cite{idrees2018composition}, and JHU-Crowd++~\cite{sindagi2020jhu} with average crowd counts of 501, 123, 815, and 346 respectively. We further benchmark on a vehicle counting dataset~\cite{Hsieh_2017_ICCV} and a penguin counting dataset~\cite{Arteta16}.

\subsubsection{Main Results.} \adrian{Our results in \cref{tab:crowd_performance} and \cref{tab:object_performance} showcase our method's performance on both crowd and vehicle counting tasks, evaluated using mean absolute error (MAE) and mean squared error (MSE). We compare our approach to unsupervised, few-shot, and zero-shot methods, where authors have made those results available. Few-shot and zero-shot methods aim to create a general counting network for any category using a large, manually annotated dataset with a diverse set of categories. Few-shot methods rely on an exemplar, sampled from the target image, to define the category of interest, whereas zero-shot methods use a text prompt. Unlike these methods, which depend on extensive annotated datasets but produce a general counting network, our unsupervised approach learns to count objects for a single category without any manual annotations. Despite this difference, we include these methods in our comparison for two reasons: First, they share the common goal of removing or reducing the annotation burden. Second, there are no unsupervised counting methods which have provided evaluation on object categories beyond crowds.}

\adrian{With respect to crowd counting, we find that our method outperforms CLIP-Count~\cite{clip_count_2023} on all available datasets. It also outperforms both CrowdCLIP~\cite{Liang_2023_CVPR} and CSS-CCNN~\cite{css_ccnn_2022} on every dataset, with the sole exception of the SHA dataset in terms of MAE. Notably, our technique significantly improves on the JHU and SHB datasets, indicating that it particularly excels in scenarios with fewer objects. With respect to object counting, there are no unsupervised counting methods which have explored objects beyond pedestrians in crowds. However, we find that our unsupervised approach surpasses existing zero-shot and few-shot methods, as demonstrated on the CARPK dataset. Additionally, we assess our method on the penguins dataset~\cite{Arteta16}, to establish a benchmark for future researchers, where we report a MAE of 5.1, and a MSE of 8.0.}

\adrian{In summary, our method not only outperforms existing unsupervised approaches in crowd counting but also outperforms zero-shot methods across both crowd and vehicle counting tasks. This suggests that our method provides superior performance for arbitrary categories.}

\subsubsection{Qualitative Results.} \adrian{We highlight qualitative results for crowd counting in \cref{fig:qual_results} by comparing count maps estimated by our method to count maps calculated from ground truth dot maps. We demonstrate that our approach produces accurate count maps that localize crowds across a wide range of counts without \textit{any} location-based supervision. We further extend our evaluation to arbitrary object categories, as seen in \cref{fig:qual_results_obj}. For penguins and vehicles, we utilize sorting data generated from training set images of the penguins~\cite{Arteta16} and the CARPK~\cite{Hsieh_2017_ICCV} datasets, respectively, assessing performance on their test sets. For flamingos, oranges, and sheep, we create synthetic images with simple prompts (e.g., ``\textit{Many sheep. Photograph.}''), combined with random scene names from the Places365 dataset~\cite{zhou2017places}. We evaluate quality on a mix of images from the FSC147 dataset~\cite{famnetcvpr2021} and manually annotated public domain images, confirming our method's robustness in accurately localizing diverse object categories.}

\subsection{Ablation Study} %
\label{sec:ablation}
\subsubsection{Impact of Partition Strategy.} In \cref{tab:partitioning_abl}, we examine the effect of partitioning strategies on network performance. Initially, we explore a straightforward approach of dividing all images into an $M\times M$ grid at different rates of $M$, which we refer to as fixed partitioning. We then evaluate a simple guided partitioning strategy, where we use the density classifier to provide a single density estimate for the whole image. If the whole image is classified as dense, we use $M\times M$ partitioning, and $1 \times 1$ otherwise. Finally, we evaluate DCGP using different partitioning rates $M$, where we start with a $1 \times 1$ rate and partition with an $M \times M$ rate for image regions determined to be dense by $h_\phi$. Notably, DCGP provides a significant performance boost compared to naively applying a $1 \times 1$ rate or $M \times M$ rate, indicating that our method is successfully addressing the challenges introduced by the synthetic label noise. Further, we determine that the optimal partitioning rate is $1 \times 1$ to $3 \times 3$, which we apply to all datasets.

\subsubsection{Impact of Pre-training.} Our approach introduces a pre-training strategy aimed at developing high-quality features for object quantity estimation, serving as a foundation for the subsequent training of our counting network. In ~\cref{tab:pre_training_abl}, we explore different pre-training strategies to determine the impact of the feature learning strategy on the final performance of the network. We compare our method with networks pre-trained on ImageNet\cite{he2016deep} and using an intra-image ranking signal~\cite{liu2018leveraging}, as well as examining the sorting signal alone. Our findings demonstrate that our pre-training technique significantly outperforms these alternatives, and that regularizing the feature space improves performance.

\subsubsection{Counting Network Training.} In \cref{sec:learning_to_count} we argued that finetuning the sorting network with the noisy synthetic counting data can distort the learned features. In \cref{tab:layer_retraining}, we examine different strategies for producing the counting network. Initially, we train the network solely with the counting signal, which confirms that relying exclusively on noisy synthetic counting data yields inferior results. Further investigation into finetuning the entire pre-trained sorting network reveals that such an approach adversely affects performance, supporting our argument that we should favor linear probing to avoid feature distortion.

\subsubsection{Impact of Image Resolution.} At inference time, we resize all images to a fixed resolution. However, many images are of a higher resolution than this fixed size. When we partition an image, we sample the patches from the higher resolution image to preserve image details. In \cref{tab:ablate_resolution}, we explore the impact of this strategy and find that it improves performance for all datasets. This effect is especially pronounced for datasets with very high resolution images like QNRF.

\section{Conclusions and Future Work}
\am{In this paper, we provide an \textit{unsupervised} counting method that can be applied to \textit{multiple object categories}.
We have tackled this challenging task by using synthetic counting data generated by latent diffusion models (LDMs), which provides a flexible strategy for effectively eliminating the annotation burden.} However, LDMs face challenges in reliably understanding object quantity, which results in noisy annotations. To mitigate this issue, we employed LDMs to create two additional types of synthetic data: one by manipulating the number of objects within images, which yields an ordered image with a weak but very reliable object quantity label, and the other by generating synthetic images with varying object density categories, offering the ability to detect dense image regions. Given these components, we deploy a novel inference time strategy that automatically detects when images are dense and then partitions those images to reduce the total number of objects per sub-image. We demonstrate our method's superiority over the SOTA in unsupervised crowd counting and zero-shot crowd counting across multiple benchmark datasets. Our work not only significantly alleviates the annotation burden but outlines a new direction for unsupervised counting. \am{Our approach, like others, has limitations. For instance, its performance falls short compared to supervised methods. Injecting limited reliable counting data into the counting network may enhance performance. Additionally, our method operates effectively on images generated by LDMs, but may not perform well on non-natural image datasets such as medical data. Addressing these challenges can be a future research direction.}

\section*{Acknowledgments}
This work was supported by the Natural Sciences and Engineering Research Council of Canada (NSERC) through the Discovery Grant program. We also extend our deepest gratitude to the Digital Research Alliance of Canada for the use of the Cedar supercomputing cluster, which was essential to our research. Finally, we thank the Computing Science Department at Simon Fraser University for their continued support and resources.

\bibliographystyle{splncs04}
\bibliography{main}

\begin{thebibliography}{10}
\providecommand{\url}[1]{\texttt{#1}}
\providecommand{\urlprefix}{URL }
\providecommand{\doi}[1]{https://doi.org/#1}

\bibitem{Arteta16}
Arteta, C., Lempitsky, V., Zisserman, A.: Counting in the wild. In: European Conference on Computer Vision (2016)

\bibitem{css_ccnn_2022}
Babu~Sam, D., Agarwalla, A., Joseph, J., Sindagi, V.A., Babu, R.V., Patel, V.M.: Completely self-supervised crowd counting via distribution matching. In: Avidan, S., Brostow, G., Ciss{\'e}, M., Farinella, G.M., Hassner, T. (eds.) Computer Vision -- ECCV 2022. pp. 186--204. Springer Nature Switzerland, Cham (2022)

\bibitem{Bai_2020_CVPR}
Bai, S., He, Z., Qiao, Y., Hu, H., Wu, W., Yan, J.: Adaptive dilated network with self-correction supervision for counting. In: Proceedings of the IEEE/CVF Conference on Computer Vision and Pattern Recognition (CVPR) (June 2020)

\bibitem{Cheng_2022_CVPR}
Cheng, Z.Q., Dai, Q., Li, H., Song, J., Wu, X., Hauptmann, A.G.: Rethinking spatial invariance of convolutional networks for object counting. In: Proceedings of the IEEE/CVF Conference on Computer Vision and Pattern Recognition. pp. 19638--19648 (June 2022)

\bibitem{cheng2022rethinking}
Cheng, Z.Q., Dai, Q., Li, H., Song, J., Wu, X., Hauptmann, A.G.: Rethinking spatial invariance of convolutional networks for object counting. In: Proceedings of the IEEE/CVF Conference on Computer Vision and Pattern Recognition. pp. 19638--19648 (2022)

\bibitem{contractor2022behavioral}
Contractor, D., McDuff, D., Haines, J.K., Lee, J., Hines, C., Hecht, B., Vincent, N., Li, H.: Behavioral use licensing for responsible ai. In: 2022 ACM Conference on Fairness, Accountability, and Transparency. pp. 778--788 (2022)

\bibitem{daless2023}
D’Alessandro, A.C., Mahdavi-Amiri, A., Hamarneh, G.: Learning-to-count by learning-to-rank. In: 2023 20th Conference on Robots and Vision (CRV). pp. 105--112 (2023). \doi{10.1109/CRV60082.2023.00021}

\bibitem{gong2022ranksim}
Gong, Y., Mori, G., Tung, F.: {R}ank{S}im: Ranking similarity regularization for deep imbalanced regression. In: International Conference on Machine Learning (ICML) (2022)

\bibitem{he2016deep}
He, K., Zhang, X., Ren, S., Sun, J.: Deep residual learning for image recognition. In: Proceedings of the IEEE conference on computer vision and pattern recognition. pp. 770--778 (2016)

\bibitem{he2023is}
He, R., Sun, S., Yu, X., Xue, C., Zhang, W., Torr, P., Bai, S., QI, X.: Is synthetic data from generative models ready for image recognition? In: The Eleventh International Conference on Learning Representations (ICLR) (2023), \url{https://openreview.net/forum?id=nUmCcZ5RKF}

\bibitem{Hsieh_2017_ICCV}
Hsieh, M.R., Lin, Y.L., Hsu, W.H.: Drone-based object counting by spatially regularized regional proposal networks. In: The IEEE International Conference on Computer Vision (ICCV). IEEE (2017)

\bibitem{hu2020count}
Hu, Y., Jiang, X., Liu, X., Zhang, B., Han, J., Cao, X., Doermann, D.: Nas-count: Counting-by-density with neural architecture search. In: Computer Vision--ECCV 2020: 16th European Conference, Glasgow, UK, August 23--28, 2020, Proceedings, Part XXII 16. pp. 747--766. Springer (2020)

\bibitem{idrees2018composition}
Idrees, H., Tayyab, M., Athrey, K., Zhang, D., Al-Maadeed, S., Rajpoot, N., Shah, M.: Composition loss for counting, density map estimation and localization in dense crowds. In: Proceedings of the European conference on computer vision (ECCV). pp. 532--546 (2018)

\bibitem{clip_count_2023}
Jiang, R., Liu, L., Chen, C.: Clip-count: Towards text-guided zero-shot object counting. In: Proceedings of the 31st ACM International Conference on Multimedia. p. 4535–4545. MM '23, Association for Computing Machinery, New York, NY, USA (2023)

\bibitem{kumar2022finetuning}
Kumar, A., Raghunathan, A., Jones, R.M., Ma, T., Liang, P.: Fine-tuning can distort pretrained features and underperform out-of-distribution. In: International Conference on Learning Representations (2022), \url{https://openreview.net/forum?id=UYneFzXSJWh}

\bibitem{Lee_2018_CVPR}
Lee, K.H., He, X., Zhang, L., Yang, L.: Cleannet: Transfer learning for scalable image classifier training with label noise. In: Proceedings of the IEEE Conference on Computer Vision and Pattern Recognition (CVPR) (June 2018)

\bibitem{lempitsky2010learning}
Lempitsky, V., Zisserman, A.: Learning to count objects in images. Advances in neural information processing systems  \textbf{23} (2010)

\bibitem{li2018csrnet}
Li, Y., Zhang, X., Chen, D.: Csrnet: Dilated convolutional neural networks for understanding the highly congested scenes. In: Proceedings of the IEEE conference on computer vision and pattern recognition. pp. 1091--1100 (2018)

\bibitem{Liang_2023_CVPR}
Liang, D., Xie, J., Zou, Z., Ye, X., Xu, W., Bai, X.: Crowdclip: Unsupervised crowd counting via vision-language model. In: Proceedings of the IEEE/CVF Conference on Computer Vision and Pattern Recognition (CVPR). pp. 2893--2903 (June 2023)

\bibitem{liu2018leveraging}
Liu, X., Van De~Weijer, J., Bagdanov, A.D.: Leveraging unlabeled data for crowd counting by learning to rank. In: Proceedings of the IEEE conference on computer vision and pattern recognition. pp. 7661--7669 (2018)

\bibitem{Paiss_2023_ICCV}
Paiss, R., Ephrat, A., Tov, O., Zada, S., Mosseri, I., Irani, M., Dekel, T.: Teaching clip to count to ten. In: Proceedings of the IEEE/CVF International Conference on Computer Vision (ICCV). pp. 3170--3180 (October 2023)

\bibitem{2020Differentiation}
Pogančić, M.V., Paulus, A., Musil, V., Martius, G., Rolinek, M.: Differentiation of blackbox combinatorial solvers. In: International Conference on Learning Representations (2020), \url{https://openreview.net/forum?id=BkevoJSYPB}

\bibitem{radford2021learning}
Radford, A., Kim, J.W., Hallacy, C., Ramesh, A., Goh, G., Agarwal, S., Sastry, G., Askell, A., Mishkin, P., Clark, J., et~al.: Learning transferable visual models from natural language supervision. In: International conference on machine learning. pp. 8748--8763. PMLR (2021)

\bibitem{famnetcvpr2021}
Ranjan, V., Sharma, U., Nguyen, T., Hoai, M.: Learning to count everything. In: Proceedings of the {IEEE/CVF} Conference on Computer Vision and Pattern Recognition (CVPR) (2021)

\bibitem{rombach2022high}
Rombach, R., Blattmann, A., Lorenz, D., Esser, P., Ommer, B.: High-resolution image synthesis with latent diffusion models. In: Proceedings of the IEEE/CVF conference on computer vision and pattern recognition. pp. 10684--10695 (2022)

\bibitem{min2022bmnet}
Shi, M., Hao, L., Feng, C., Liu, C., Cao, Z.: Represent, compare, and learn: A similarity-aware framework for class-agnostic counting. In: Proc. IEEE/CVF Conference on Computer Vision and Pattern Recognition (CVPR) (2022)

\bibitem{Shipard_2023_CVPR}
Shipard, J., Wiliem, A., Thanh, K.N., Xiang, W., Fookes, C.: Diversity is definitely needed: Improving model-agnostic zero-shot classification via stable diffusion. In: Proceedings of the IEEE/CVF Conference on Computer Vision and Pattern Recognition (CVPR) Workshops. pp. 769--778 (June 2023)

\bibitem{sindagi2019pushing}
Sindagi, V.A., Yasarla, R., Patel, V.M.: Pushing the frontiers of unconstrained crowd counting: New dataset and benchmark method. In: Proceedings of the IEEE International Conference on Computer Vision. pp. 1221--1231 (2019)

\bibitem{sindagi2020jhu}
Sindagi, V.A., Yasarla, R., Patel, V.M.: Jhu-crowd++: Large-scale crowd counting dataset and a benchmark method. IEEE Transactions on Pattern Analysis and Machine Intelligence  \textbf{44}(5),  2594--2609 (2020)

\bibitem{trivedi2023a}
Trivedi, P., Koutra, D., Thiagarajan, J.J.: A closer look at model adaptation using feature distortion and simplicity bias. In: The Eleventh International Conference on Learning Representations (2023), \url{https://openreview.net/forum?id=wkg_b4-IwTZ}

\bibitem{wan2021generalized}
Wan, J., Liu, Z., Chan, A.B.: A generalized loss function for crowd counting and localization. In: Proceedings of the IEEE/CVF Conference on Computer Vision and Pattern Recognition. pp. 1974--1983 (2021)

\bibitem{wang2020distribution}
Wang, B., Liu, H., Samaras, D., Nguyen, M.H.: Distribution matching for crowd counting. Advances in neural information processing systems  \textbf{33},  1595--1607 (2020)

\bibitem{gao2020nwpu}
Wang, Q., Gao, J., Lin, W., Li, X.: Nwpu-crowd: A large-scale benchmark for crowd counting and localization. IEEE Transactions on Pattern Analysis and Machine Intelligence  (2020). \doi{10.1109/TPAMI.2020.3013269}

\bibitem{NEURIPS2020_f4e3ce3e}
Wu, P., Zheng, S., Goswami, M., Metaxas, D., Chen, C.: A topological filter for learning with label noise. In: Larochelle, H., Ranzato, M., Hadsell, R., Balcan, M., Lin, H. (eds.) Advances in Neural Information Processing Systems. vol.~33, pp. 21382--21393. Curran Associates, Inc. (2020), \url{https://proceedings.neurips.cc/paper_files/paper/2020/file/f4e3ce3e7b581ff32e40968298ba013d-Paper.pdf}

\bibitem{zhang2016single}
Zhang, Y., Zhou, D., Chen, S., Gao, S., Ma, Y.: Single-image crowd counting via multi-column convolutional neural network. In: Proceedings of the IEEE conference on computer vision and pattern recognition. pp. 589--597 (2016)

\bibitem{zhou2017places}
Zhou, B., Lapedriza, A., Khosla, A., Oliva, A., Torralba, A.: Places: A 10 million image database for scene recognition. IEEE Transactions on Pattern Analysis and Machine Intelligence  (2017)

\bibitem{pmlr-v162-zhu22a}
Zhu, Z., Dong, Z., Liu, Y.: Detecting corrupted labels without training a model to predict. In: Proceedings of the 39th International Conference on Machine Learning. Proceedings of Machine Learning Research, vol.~162, pp. 27412--27427. PMLR (17--23 Jul 2022), \url{https://proceedings.mlr.press/v162/zhu22a.html}

\end{thebibliography}
\clearpage
\appendix
\section*{\LARGE Supplementary Materials}

\section{Evaluating Synthetic Dataset Quality}
Our goal is to assess the quality of the synthetic datasets we use to train our unsupervised counting model. This section provides a thorough analysis of these datasets, aiming to deepen the understanding of our method's strengths and limitations.

\subsection{Synthetic Counting Data}
\label{sec:syn_cnt_data}
\begin{table}[b]
  \caption{\textbf{Prompt Count Noise.} we manually annotate 40 synthetic examples for each of several prompt count categories. We evaluate the statistics of the true underlying counts for each prompt count category. }
  \label{tab:prompt_noise}
  \centering
    \begin{tabular}{ccccc}
    \hline
    Prompt Count & Mean & Std. & MAE & rMAE \\
    \hline
    1 & 1.02 & 0.34 & 0.07 & 0.07 \\
    5 & 4.26 & 0.79 & 0.88 & 0.18 \\
    10 & 10.0 & 3.48 & 2.57 & 0.26 \\
    15 & 14.86 & 5.73 & 4.86 & 0.32 \\
    20 & 23.07 & 11.10 & 8.98 & 0.45 \\
    40 & 49.29 & 39.00 & 24.38 & 0.61 \\
  \bottomrule
  \end{tabular}
\end{table}
\subsubsection{Understanding Prompt Label Noise.} In the beginning of our paper, we introduced the concept of label noise when using a latent diffusion model (LDM), such as Stable Diffusion, to generate images with a specified number of objects. Often, the actual number of objects in these images doesn't align with the requested counts. To understand this discrepancy better, we manually annotated 40 synthetic examples for counts ranging from 1 to 40. Our in-depth analysis, shown in \cref{tab:prompt_noise}, highlights how the error between the requested and actual counts grows with the increase in the prompt count. Although these findings may not be universally applicable across all ranges or object categories, they provide useful insights into how synthetic counting data affected by label noise can still be valuable for learning. Notably, we observed that the average for the true underlying counts is often close to the requested count, despite significant variations. Often, the average count for these distributions is within 15\% of the prompt label, indicating that Stable Diffusion produces a distribution of images with true counts centered near the desired amounts

\subsubsection{Impact of Outliers.}
\begin{table}[th]
  \caption{\textbf{Outlier Removal Ablation Study.} We explore the performance implications of filtering likely outliers in the noisy synthetic counting dataset.}
  \label{tab:outlier_removal}
  \centering
    \begin{tabular}{lcccccccc}
    \hline
    \multicolumn{1}{c}{}&  \multicolumn{2}{c}{SHB}  & \multicolumn{2}{c}{JHU} &  \multicolumn{2}{c}{SHA} &  \multicolumn{2}{c}{QNRF} \\
    \cline{2-9}
    Strategy & \multicolumn{1}{c}{MAE} & \multicolumn{1}{c}{MSE} & MAE & MSE & MAE & MSE & MAE & MSE \\
    \hline
    No Outlier Removal & 48.2 & 67.2 &  189.0 & 582.4  & 188.3 & 292.7 & 366.8 & 558.3  \\
    Outlier Removal & \textbf{35.0} & \textbf{50.7} &  \textbf{173.8} & \textbf{519.4}  & \textbf{152.7} & \textbf{219.0} & \textbf{283.1} & \textbf{453.2}  \\
  \bottomrule
  \end{tabular}
\end{table}
Within our methodology, we introduced a straightforward outlier removal technique to help reduce label noise. \cref{tab:outlier_removal} evaluates the effectiveness of this method in the context of crowd counting. The results indicate that this basic form of outlier removal significantly boosts performance across different crowd counting datasets, highlighting its importance in enhancing model accuracy. These results suggest that label noise hampers performance, especially in datasets with larger average counts, such as SHA and QNRF. Nonetheless, our analysis also indicates that identifying and mitigating this noise is feasible, and further investigation into noise reduction strategies could be highly advantageous.

\subsubsection{Impact of the Maximum Prompt Count.}
\begin{figure}[tb]
  \centering
  \includegraphics[width=0.8\textwidth]{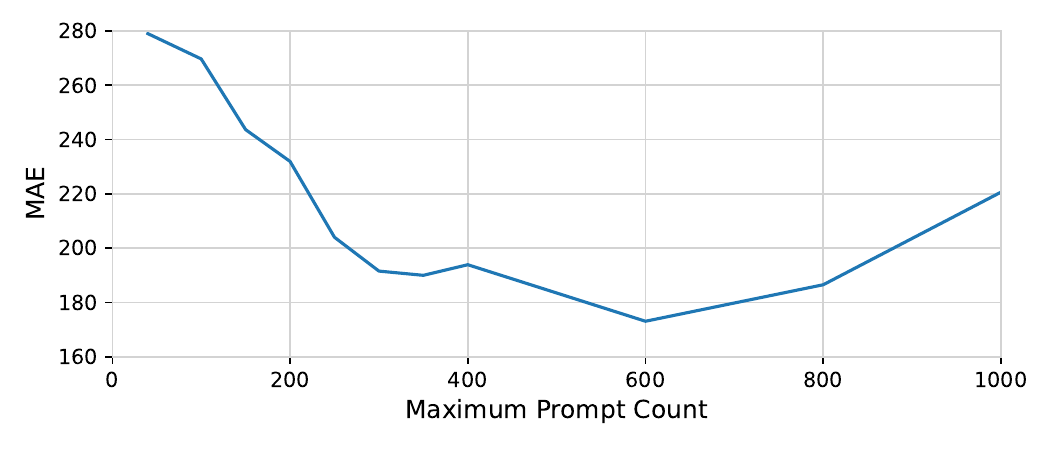}
  \caption{\textbf{Impact of Maximum Prompt Count}. We overview the performance impact of changing the maximum prompt count used when training the counting network. Our method is remains accurate across a range of values.}
  \label{fig:max_prompt_count}
\end{figure}
In our approach to generating noisy synthetic counting data, we employ a wide range of prompt counts to create a rich dataset. This naturally raises the question of whether the range of counts used during training influences the training and performance of our counting network.  Specifically, in \cref{fig:max_prompt_count}, we explore the effect of setting different maximum prompt counts ($c_{max}$) on the model's accuracy using the JHU++ test set. This involves using all synthetic counting images with a prompt count equal to or lower than $c_{max}$. Our analysis reveals an optimal range for this maximum count—specifically, between 250 and 800 for the JHU dataset. This optimal range underscores our method's adaptability, showing it can handle a wide variety of scenarios effectively.

\begin{figure}[tb]
  \centering
    \includegraphics[width=\textwidth]{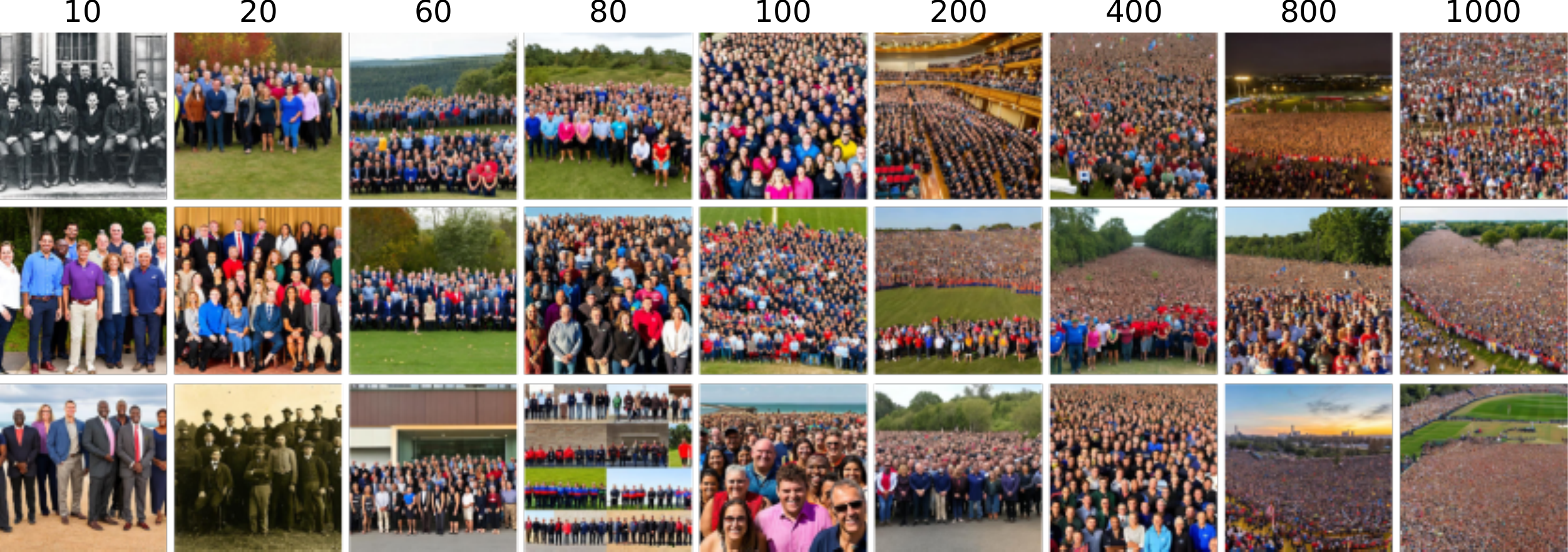}
    \caption{\textbf{Synthetic Crowd Count Qualitative.} We highlight samples of synthetic counting images from a wide range of prompt count categories. We demonstrate that images contain realistic crowds which are organized in a natural way.}
    \label{fig:syn_cnt_qual}
\end{figure}

\begin{figure}[tb]
  \centering
    \includegraphics[width=\textwidth]{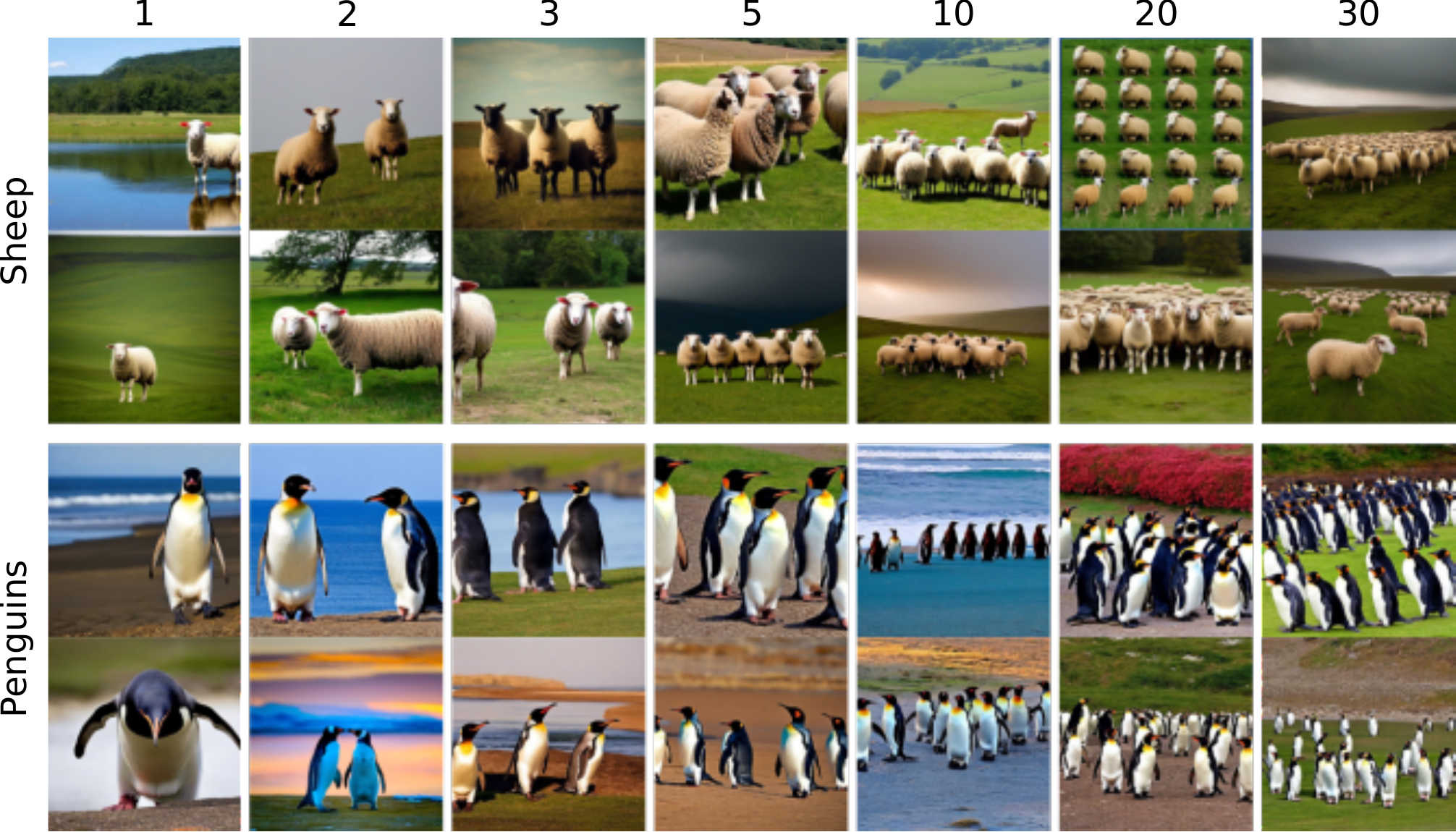}
    \caption{\textbf{Synthetic Object Count Qualitative.} Stable Diffusion can produce synthetic counting data for a wide range of objects, including sheep and penguins.}
    \label{fig:syn_sheep_qual}
\end{figure}
\subsubsection{Qualitative.} In \cref{fig:syn_cnt_qual}, we provide synthetic counting samples for a crowd counting problem across a wide range of crowd densities. Despite the presence of label noise, synthetic images remain reasonably consistent within expected ranges, even at high prompt counts. For instance, images prompted with a count of 1000 exhibit significant noise, often featuring several thousand objects. Nonetheless, these images consistently depict large, dense crowds, suggesting that Stable Diffusion retains an understanding of quantity to some extent, even at these higher numbers. This observation aligns with our findings in \cref{sec:syn_cnt_data} and \cref{tab:prompt_noise}, where the average true count if often similar to the prompted count.

Moreover, our analysis sheds light on scene biases within the synthetic counting images. Predominantly, the images showcase clear, daylight settings, frequently set in expansive outdoor areas with greenery. Despite this tendency, the dataset also exhibits variety; some images mimic historical photographs, while others depict indoor scenes, such as concert halls, indicating a diversity in the visual contexts of the generated crowds.

\subsection{Synthetic Sorting Data} 
\begin{table}[b]
  \caption{\textbf{Object Removal.} We manually annotate 50 synthetic sorting examples from each dataset to determine how frequently Stable Diffusion successfully removes objects from a reference image. We corroborate this by using a fully-supervised crowd counting model (DM-Count~\cite{wang2020distribution}) to estimate the synthetic image counts for \textit{all} examples.}
  \label{tab:object_removal}
  \centering
    \begin{tabular}{ccccc}
    \hline
    Estimate Type & SHB & JHU & SHA & QNRF \\
    \hline
    Manual & 100.0\% & 96.0\% & 98.0\% & 96.0\% \\
    DM-Count~\cite{wang2020distribution} & 99.6\% & 90.2\% & 99.2\% & 97.2\% \\
  \bottomrule
  \end{tabular}
\end{table}
\subsubsection{Removal Accuracy.} To substantiate the reliability of the synthetic sorting data, we examine 50 object removal examples from each dataset. We only analyze object removal, due to the fact that outpainting inherently preserves original objects, and thus these images always have at least as many objects as the reference image. This inspection revealed minimal discrepancies: ShanghaiTechB produced no incorrectly ranked examples, while ShanghaiTechA exhibited only one. QNRF and JHU++ presented slightly higher instances of two incorrectly ranked examples each. 

Moreover, to corroborate our manual assessment, we employed a fully supervised DM-COUNT~\cite{wang2020distribution} model to evaluate crowd count estimations across 500 real and synthetic removal pairs from each dataset. The resultant accuracy rates were high: 99.6\% for ShanghaiTechB, 99.2\% for ShanghaiTechA, 97.2\% for QNRF, and 90.2\% for JHU++. However, these results are dependent on the accuracy of the model, and are only meant to compliment the manually collected annotations. These findings, summarized in \cref{tab:object_removal}, affirm the credibility of the synthetic object removal strategy used during the data generation process.

\subsection{Synthetic Density Classification Data} 
\begin{figure}[tb]
  \centering
  \includegraphics[width=\textwidth]{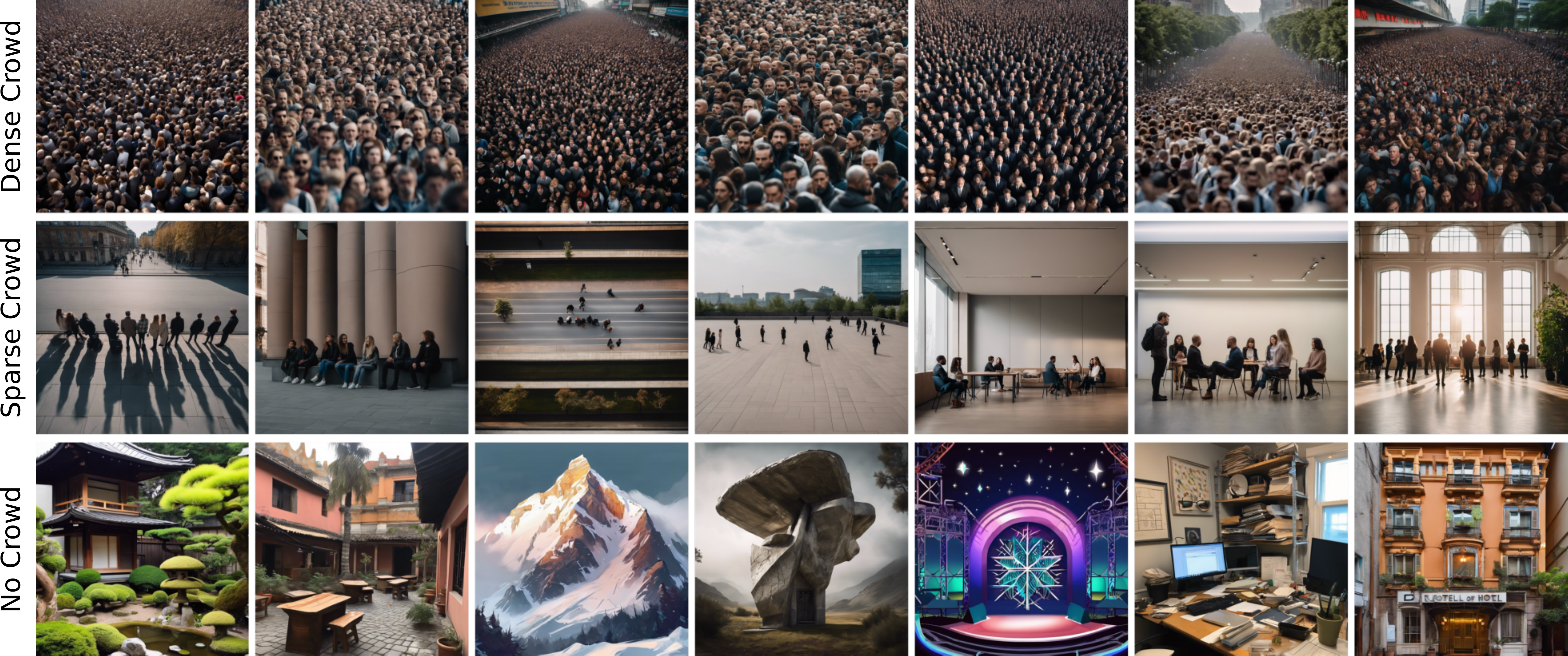}
  \caption{\textbf{Synthetic Density Qualitative}. Stable Diffusion excels in producing diverse and high-quality images across various density categories, encompassing dense, sparse, and empty scenes with broad coverage.}
  \label{fig:density_samples}
\end{figure}
\begin{figure}[tb]
  \centering
  \includegraphics[width=\textwidth]{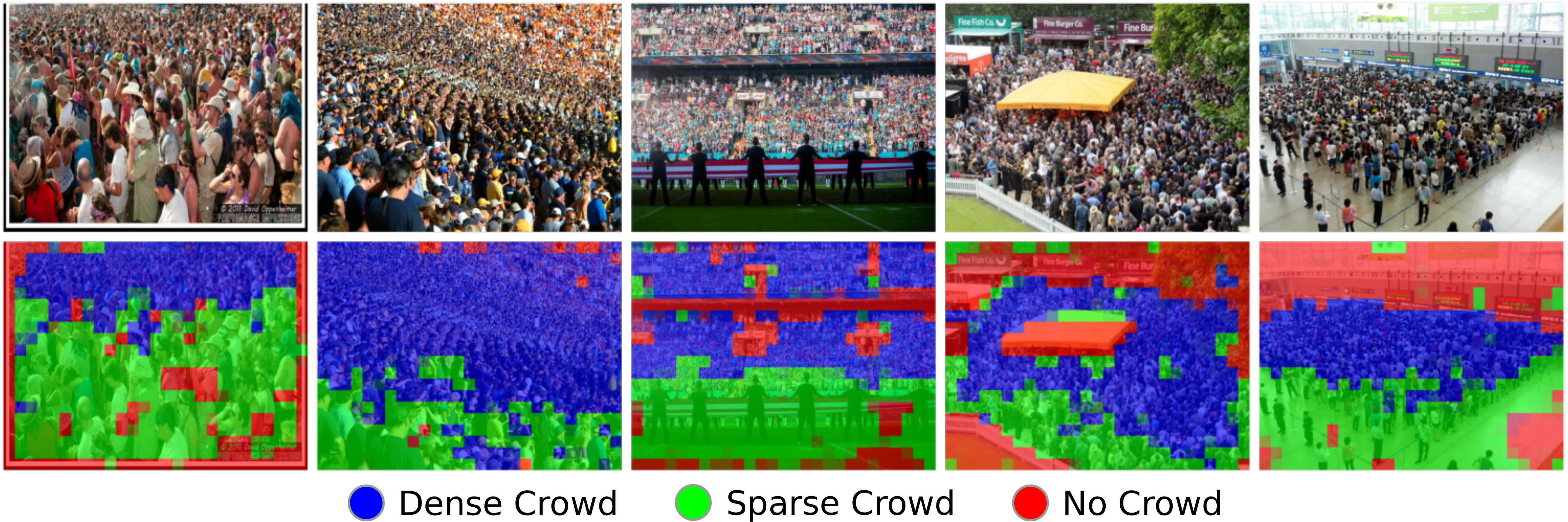}
  \caption{\textbf{Density Class Maps}. We overview the class maps generated by passing the sorting network features through the density classifier. Our method does not utilize any localization information, and yet it accurately localizes crowded regions within images.}
  \label{fig:density_class_map}
\end{figure}
We observe that Stable Diffusion excels in creating density classification data, which plays a crucial role in training our density classification network. In \cref{fig:density_samples}, we showcase synthetic images crafted specifically for crowd counting scenarios. These examples demonstrate Stable Diffusion's capability to generate diverse and accurate representations for such tasks.

\cref{fig:density_class_map} presents the classification maps generated by our network, trained on this synthetic data. The maps reveal the network's proficiency in accurately identifying areas with dense and sparse crowds, as well as empty spaces devoid of pedestrians. This underscores the significant impact and utility of utilizing synthetic data for density classification

\section{Expanded Qualitative Analysis}

\begin{figure}[!h]
  \centering
  \includegraphics[width=\textwidth]{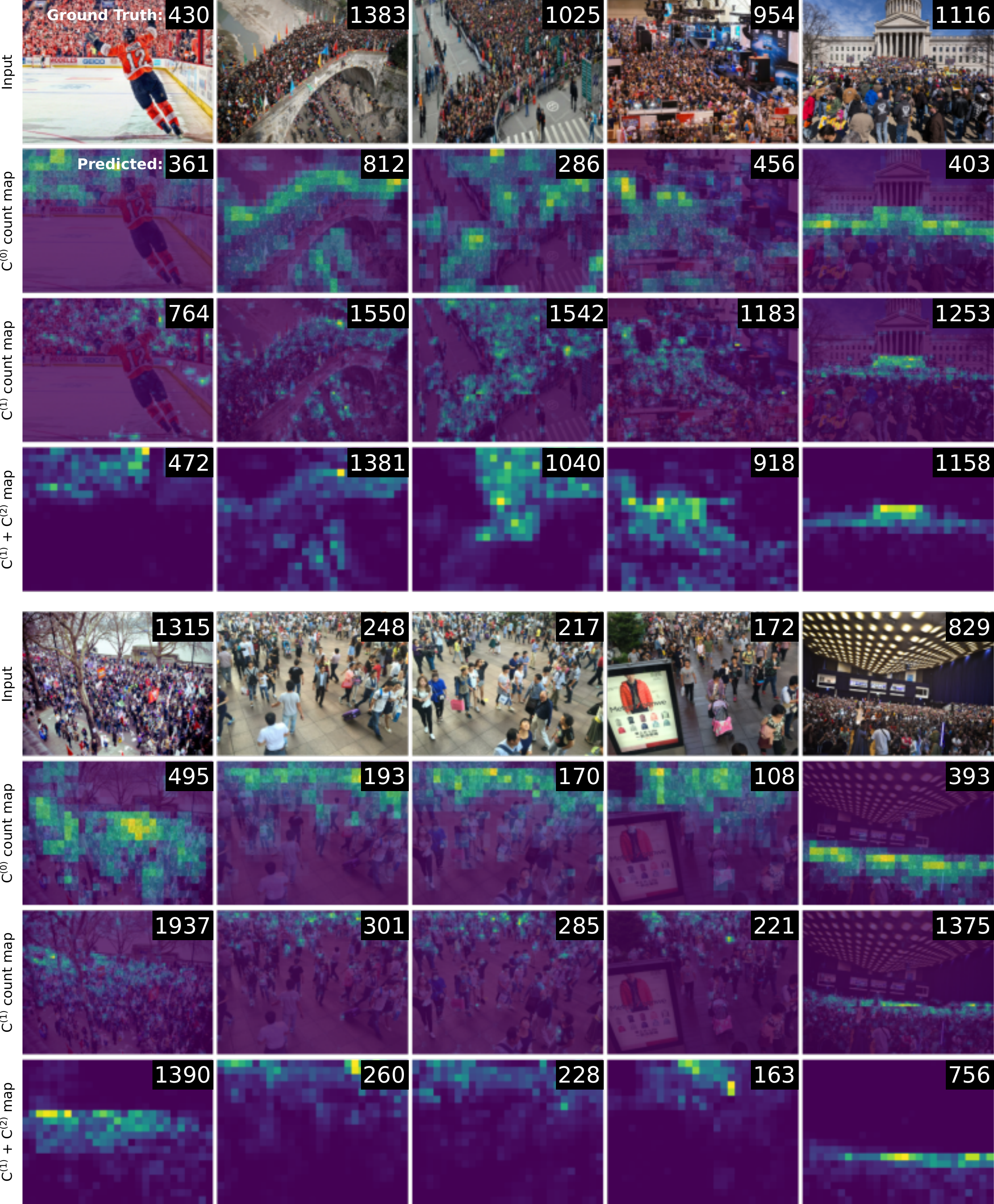}
  \caption{\textbf{Crowd Counting Qualitative}. We explore the quality of the model output for the $c^{(0)}$, $c^{(1)}$, and density guided count maps which join $c^{(0)}$, $c^{(1)}$. Input images are annotated with the ground truth count, and count maps are annotated with the estimated count for that map.}
  \vspace{4em}
  \label{fig:extended_crowd_qualitative}
\end{figure}
\subsubsection{Crowd Counting.}
In \cref{fig:extended_crowd_qualitative}, we delve deeper into the crowd counting challenge with an expanded qualitative analysis. This section sheds light on our method's process, which integrates whole image estimates with those from densely populated patches identified within the image. We illustrate this approach by presenting the initial count map estimate, $C^{(0)}$, for the entire image, alongside $C^{(1)}$, which represents the concatenated estimates for partitioned patches from a $3\times3$ grid. Additionally, we show the combined count map resulting from our density classifier guided partitioning technique. These examples are valuable for understanding how our model calculates various estimates to achieve a precise overall count. Moreover, they demonstrate our method's proficiency in accurately identifying crowd locations, whether analyzing the complete image or specific patches. Importantly, our model proves resilient against a wide array of non-target elements within a scene, such as buildings, roads, and natural features, underscoring its robustness in complex environments.
\subsubsection{Extension to Similar Objects.}
\begin{figure}[tb]
  \centering
    \includegraphics[width=\textwidth]{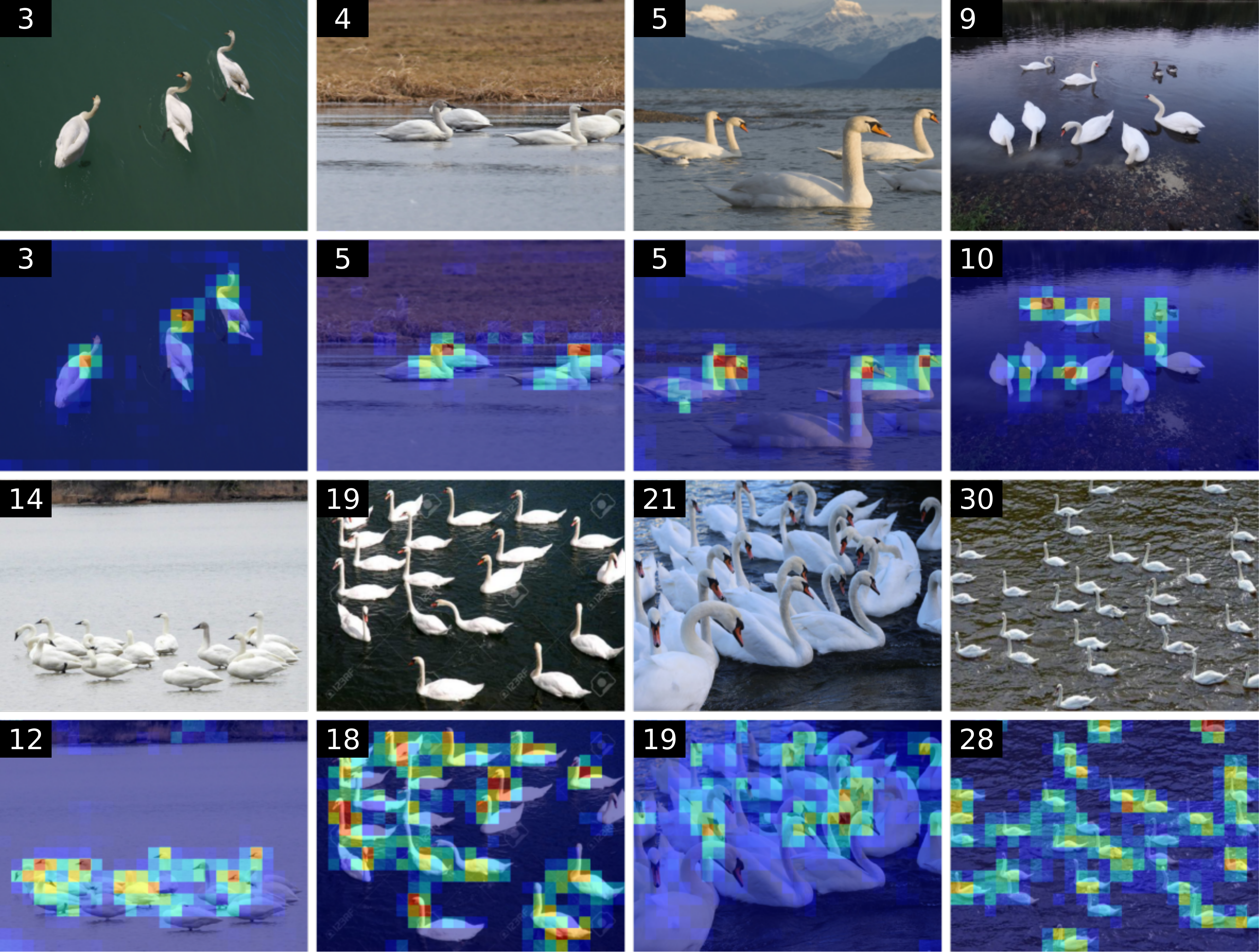}
    \caption{\textbf{Extension to Similar Objects.} We investigate an emergent capability of our system to generalize to similar object categories. For example, a model trained to identify flamingos is also capable of identifying waterfowl.}
    \label{fig:flamingo_to_swan}
\end{figure}
In this section, we examine our model's ability to recognize objects outside its training set but within related categories. \cref{fig:flamingo_to_swan} demonstrates that a model trained on flamingos can also accurately identify waterfowl, showcasing impressive generalization to similar objects. This versatility is significant, indicating that a model trained on a specific object category can be effectively applied to analogous categories, enhancing its utility and broadening its applicability.
\section{Implementation Details}
\label{sec:details}
\subsubsection{Implementation.}
\label{sec:implementation}
We employ ResNet50~\cite{he2016deep} as the underlying architecture. We train $f_\theta$ for 5 epochs, utilizing the Adam optimizer with a learning rate set to $5e^{-5}$. We resize all images to a uniform size of (640, 853, 3). During inference, we use a partition rate of 3 for all datasets. For the data generation process, we rely on Stable Diffusion 2.1. When performing image-to-image generation, we set the strength parameter to 0.45. Throughout all image generation procedures, we maintain a fixed guidance scale of 7.5 and carry out optimization for 50 steps. For crowd counting problems, we set the prompt labels to: $$ N = [0, 1, 5, 10, 15, 20, 40, 60, 80, 100, 150, 200, 250, 300, 350, 400, 600, 800, 1000].$$ And for all other object categories, which have significantly lower average counts, we set the prompt labels to:
$$ N = [0, 1, 2, 3, 5, 10, 15, 20, 30, 60].$$
Further, we set aside 15\% of the synthesized sorting data as a validation set for performing count model selection. We use this data to set the maximum prompt count, $c_{max}$, in the set $N$ when training the counting network $g_\Phi$. We do this by performing inference on these validation sorting examples with $g_\Phi$ and selecting the model with the highest accuracy. This provides a $c_{max}$ of 150 for SHB, 600 for SHA, 600 for JHU, and 1000 for QNRF, which approximately follows the mean of each dataset.
\subsubsection{Prompt Selection.}
\begin{table}[h]
  \caption{\textbf{Prompt List.} This list showcases the straightforward prompts utilized for data generation, emphasizing the simplicity of the generation process.}
  \label{tab:prompt_list}
  \centering
    \begin{tabular}{ll|p{4.5cm}|p{4.5cm}}
    \hline
    Category & Usecase & Prompt & Negative \\
    \hline
    \rowcolor{Gray}Crowd & Count & A group of $\{N\}$ people. & - \\
    Vehicle & Count & $\{N\}$ vehicles. Overhead view. & - \\
    \rowcolor{Gray}Penguin & Count & $\{N\}$ Penguins. & - \\
    
    Crowd & Remove & An empty outside space. Nobody around. & people. crowds. pedestrians. humans. \\
    \rowcolor{Gray}Penguins & Remove & An empty outdoor arctic space. Nobody around. & penguins, birds, animals, fowl, avian. \\
    Vehicles & Remove & An empty parking lot. Nobody around. & vehicles, cars, automobiles, trucks, jeeps, suvs, vans. \\
    \rowcolor{Gray} Crowd & Add & A crowd of people. & - \\
    Penguins & Add & A large group of penguins. & - \\
    \rowcolor{Gray}Vehicles & Add & A busy parking lot with many cars. Overhead view. & - \\
  \bottomrule
  \end{tabular}
\end{table}
In \cref{tab:prompt_list}, we present a selection of prompts utilized for various synthesis tasks and object categories. It is not meant to be an exhaustive list, but rather to highlight the simplicity of the prompts used. The most complex specification involves directing that vehicle counting data be generated from an overhead perspective to better align the synthetic images with the aerial drone photography found within the CARPK dataset. All datasets utilize the following negative prompt list to ensure realism: \textit{artistic, painting, vector art, graphic design, watercolor, text, writing, anime}.
\subsubsection{Inference Throughput.} This section assesses the inference speed of our model, conducted on a Nvidia Titan X Pascal GPU. Our findings reveal that the model processes images with dimensions (640, 853, 3) at an average rate of 49.5 frames per second (FPS). However, the model's speed varies when handling images with high-density areas, necessitating subdivision into sub-patches for detailed analysis. Specifically, in scenarios requiring the image to be divided into a $3 \times 3$ grid due to dense regions, the throughput decreases to an average of 5.5 FPS. This variation outlines the range of our model's inference speed, providing insights into its performance under different conditions.
\section{Failure Cases}
\begin{figure}[tb]
  \centering
  \includegraphics[width=\textwidth]{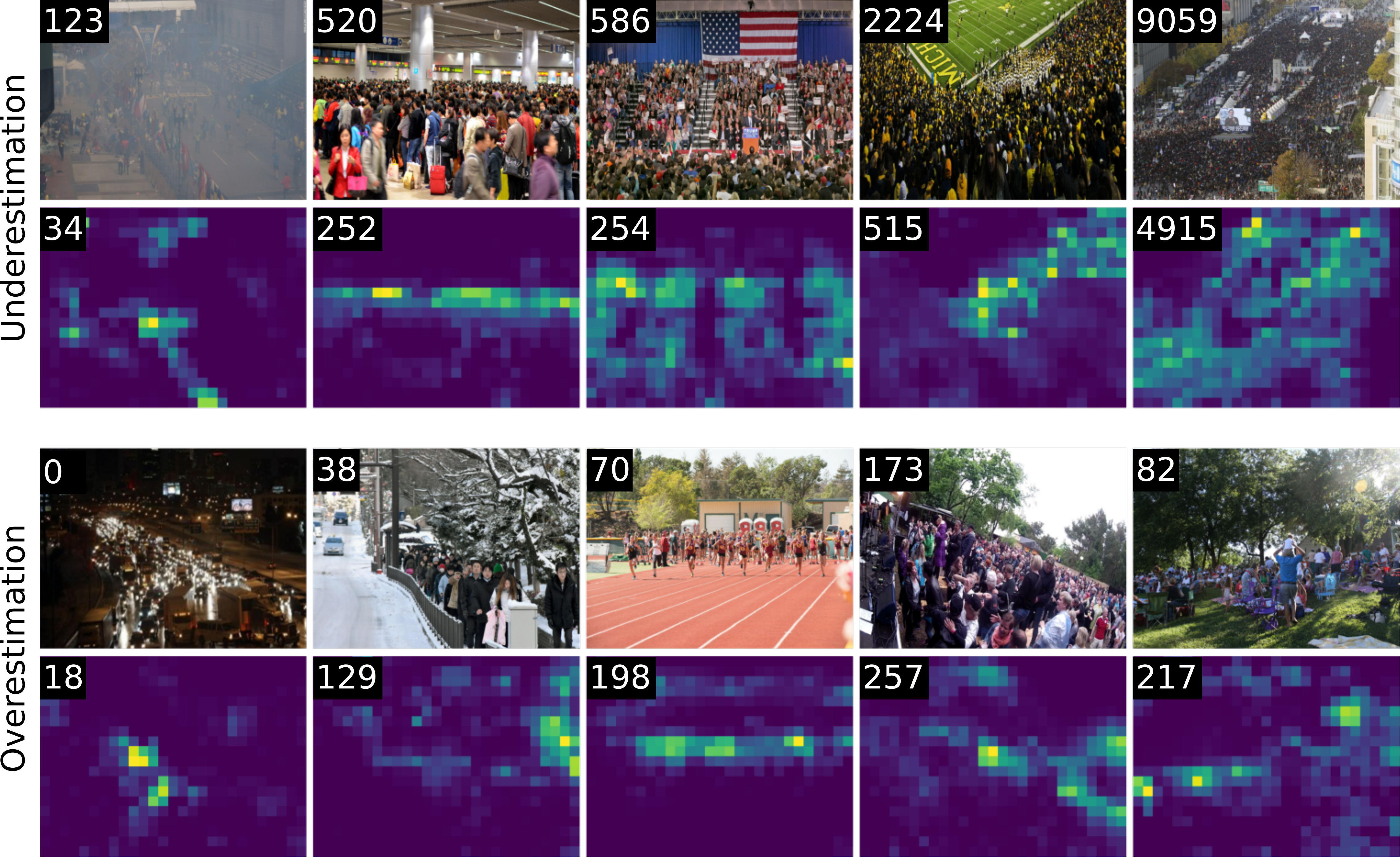}
  \caption{\textbf{Failure Cases}. We highlight instances of underestimation and overestimation by our model, including challenges posed by environmental factors like fog and clutter, and the model's underestimation in some dense scenes.}
  \label{fig:failure_cases}
\end{figure}
In \cref{fig:failure_cases}, we highlight instances where our model does not perform as expected, categorizing these into underestimation and overestimation failures. Underestimation can occur for several reasons. For instance, in the top left-most image, fog obscures pedestrians, making it difficult for our model to recognize them. Similarly, in other cases, people are hidden behind one another, or there simply isn't enough visible detail due to the object size or image resolution for accurate detection. On the other hand, overestimation issues often arise from the model mistaking unrelated elements for the target object. An example includes a cluttered traffic scene at night being misinterpreted as a group of pedestrians. Similarly, image regions with trees and dense foliage sometimes confuse the model. Despite these challenges, our method generally performs well across a range of scenarios. Nevertheless, these limitations highlight areas for improvement in future research.
\section{Negative Social Impacts \& Human Subjects Data}
The modern deep learning approaches to crowd counting emerged with a key paper published in 2010~\cite{lempitsky2010learning}. This area of research, crucial for tasks like event management, disaster response, and public safety enhancement, has seen substantial developments over the years. For instance, advanced crowd counting techniques played a pivotal role in analyzing crowd behavior during significant events, such as the January 6th Capitol riot~\cite{Cheng_2022_CVPR}, which underscored its societal importance. However, key datasets like ShanghaiTech A and B~\cite{zhang2016single}, JHU++~\cite{sindagi2019pushing, sindagi2020jhu}, QNRF~\cite{idrees2018composition}, and NWPU~\cite{gao2020nwpu} have relied heavily on images from public surveillance and the internet. This raises concerns about privacy and the potential for misuse in surveillance by various entities.

it is important to note that crowd counting datasets do not contain information related to facial recognition or individual identification; they merely mark the location of persons with dots or bounding boxes without revealing any personal details which somewhat mitigates privacy concerns. However, concerns remain since individuals might unknowingly appear in these datasets. Some dataset creators, like the authors of JHU++, offer a removal process for those depicted in images, but this process often lacks transparency. We advocate for clearer and more efficient opt-out procedures to protect individual privacy.

Moreover, the potential misuse of crowd counting in surveillance applications cannot be overlooked. Although crowd counting is distinct from facial recognition and not intended for invasive monitoring, its misuse remains a concern. Dataset licenses usually restrict use to academic and non-commercial purposes, yet these licenses may still be too permissive to prevent downstream harms. Recent proposals like the Open Responsible AI License (OpenRAIL)~\cite{contractor2022behavioral} have been introduced to ensure AI's ethical use, especially concerning applications that could infringe on personal rights or safety. We propose that these more restrictive licenses should be more widely adopted in the field.

We argue that the potential of crowd counting methods to serve societal good outweighs the limited scope for misuse. Nevertheless, the ethical implications of these technologies demand continuous vigilance from researchers. It is imperative for contributors in this domain to be conscientious of how and by whom their work is utilized, maintaining an awareness of the broader societal implications of their contributions.

\end{document}